\documentclass[11pt]{article}

\usepackage[preprint]{acl}

\usepackage{times}
\usepackage{latexsym}

\usepackage[T1]{fontenc}

\usepackage[utf8]{inputenc}

\usepackage{microtype}

\usepackage{inconsolata}

\usepackage{graphicx}
\usepackage{booktabs}
\usepackage[table]{xcolor}
\usepackage{amsmath}
\usepackage{amssymb}
\usepackage{mathtools}
\usepackage{amsthm}
\usepackage{comment}
\usepackage[capitalize,noabbrev]{cleveref}

\usepackage{microtype}
\usepackage{graphicx}
\usepackage{subcaption}
\usepackage{enumitem}
\usepackage{multirow}
\usepackage{fancyvrb}
\usepackage{listings}
\usepackage{makecell}  
\usepackage{xcolor}  
\usepackage{longtable}  
\usepackage{pifont}
\usepackage{wrapfig}
\usepackage{array}
\usepackage{caption}
\usepackage{tcolorbox}
\usepackage{tikz}
\usetikzlibrary{positioning}
\tcbuselibrary{skins,breakable}

\title{Anticipatory Evaluation of Language Models: \\ Predicting LLM Benchmark Scores from Descriptions}

\author{Jungsoo Park$^{1}$\quad Ethan Mendes$^{1}$\quad Gabriel Stanovsky$^{2}$\quad Alan Ritter$^{1}$ \\
  Georgia Institute of Technology$^{1}$ \quad The Hebrew University of Jerusalem$^{2}$\\
  \texttt{jpark3272@gatech.edu} \quad \texttt{emendes3@gatech.edu} \\
  \texttt{gabriel.stanovsky@mail.huji.ac.il} \quad
  \texttt{alan.ritter@cc.gatech.edu} \\}

\newcommand
{
\datasetname
}{
\textsc
{EvalCast}}

\begin{document}
\maketitle
\begin{abstract}

Modern ML research increasingly faces an evaluation bottleneck: building benchmarks and running experiments is costly, slowing iteration.
We ask how much can be inferred about evaluation outcomes before experiments, given only a natural-language description of the task and setup.
We call this setting \emph{anticipatory evaluation}.
We introduce \datasetname, a literature-grounded resource of 2,290 description–performance pairs from 1,519 arXiv papers, including a test set of papers published after the model’s knowledge cutoff.
We evaluate two tasks: predicting absolute performance through regression and comparing benchmark difficulty through pairwise prediction.
Experiments with frontier LLMs show that accurate regression remains challenging, while pairwise predictions can recover relative performance ordering under matched settings.
Comparable performance on post-cutoff and streaming evaluations suggests that the observed signal is not solely attributable to previously reported results.
Overall, \datasetname~provides a resource and evaluation framework for studying anticipatory evaluation and its potential use in prioritizing benchmarks.

\end{abstract}

\section{Introduction}


A growing fraction of machine learning progress is constrained not only by model development, but also by the cost of \emph{evaluation}.
Building a new benchmark, annotating data, and comparing model variants can require substantial time, engineering effort, and compute.
Researchers may incur these costs before knowing whether the proposed task provides a meaningful evaluation.
For instance, the task may be too easy, too hard, or insufficiently discriminative among the models being evaluated.
What if it was possible to know how difficult a proposed task is before building a benchmark?

Motivated by this challenge, we ask how much can be inferred about expected benchmark performance from only a natural-language description of the task and experimental setup. 
We call this setting \emph{anticipatory evaluation}.
Our goal is not to replace empirical benchmarking, but to study signals available before any benchmark is available for prioritization.
This is analogous to virtual screening in drug discovery, where scoring functions help to prioritize candidate molecules for expensive experimentation~\citep{sanchez2018inverse,han2025aidriven}.
Researchers similarly face many candidate tasks, dataset variants, and evaluation protocols, but can empirically explore only a subset.
For example, a researcher seeking to identify model weaknesses could use anticipatory forecasts to prioritize benchmarks where the model is expected to struggle.


This framing differs from the existing literature on performance prediction.
Prior work has inferred missing results from observed experimental tables~\citep{xia2020predicting, ye2021finegrained, ye2023predictable, zhaocan} and estimated full-benchmark performance from small pilot runs~\citep{maiapolo2024tinybenchmarks, pacchiardi2024hundred, anugraha2025proxylm}.
Other work has modeled performance trends across scale, data, and compute~\citep{kaplan2020scaling, hoffmann2022training}.
More recently, \citet{wen2025predicting} explored predicting which of two methods would perform better on existing benchmarks.
In contrast, we predict a fixed model’s performance on novel benchmarks from their descriptions alone, without requiring empirical measurements.
This enables early screening of candidate benchmarks.


To study this setting, we introduce \datasetname, a literature-grounded resource of description–performance pairs curated from arXiv papers.
The dataset spans diverse tasks, domains, and metrics, with splits that carefully control for data contamination.
We evaluate anticipatory prediction using two tasks: a regression task that measures mean absolute prediction error, in addition to a pairwise evaluation task that measures the ability of models to rank evaluation configurations by difficulty.
Our results show that accurate regression remains
challenging, while pairwise prediction reveals
a useful signal for prioritizing candidate evaluations.
Controlled ablations and error analysis further characterize which task characteristics affect predicted difficulty and when predictions fail.

Taken together, these results suggest that anticipatory evaluation can provide a coarse signal for screening and prioritization.  While prediction is not a substitute for empirical evaluation, we argue that this approach could be useful in the design of future benchmarks.
Our main contribution is to establish anticipatory evaluation as an empirical research problem and provide a resource and evaluation methodology for studying it systematically.
Our experiments with frontier LLMs characterize current capabilities and limitations, providing baselines for future work.
\footnote{Our data and code are available in \url{https://github.com/JJumSSu/EVALCAST}}
\section{Related Work}

\paragraph{Performance Prediction}

Prior works primarily fall into two categories.
(1) \emph{Missing-entry prediction}: infer unobserved cells from observed ones using matrix-completion~\citep{achille2019task2vec,xia2020predicting,chen2021model,ye2021finegrained,ye2023predictable,rabinovich2023predicting,shi2025importance,zhaocan}; these assume seed results, stable task identities, and overlap for interpolation.
(2) \emph{Efficient evaluation}: run a small subset of trials and extrapolate full-benchmark performance~\citep{maiapolo2024tinybenchmarks,pacchiardi2024hundred,anugraha2025proxylm}; these still require executing target (or closely related) runs.
Finally, we note work by \citet{wen2025predicting}, which aims to predict which of two novel research ideas is likely to perform better on a set of {\em existing benchmarks}. In contrast, we study whether the performance of a fixed model on a {\em novel benchmark} can be anticipated from its description alone (see Table~\ref{tab:difference}).

\paragraph{LLMs for Prediction}
LLMs have been used as regressors in AutoML and black-box optimization~\citep{song2024omnipred,vacareanu2024words,nguyen2024predicting,requeima2024llm,tang2024understanding}, often with structured inputs, seed trials, or target-task finetuning.
In contrast, we predict performance from free-form descriptions of novel datasets and evaluation protocols, without running the target model on those datasets~\citep{zou2022forecasting,schoenegger2023large,kargerforecastbench,halawi2024approaching,abolghasemi2025humans,moussa2025scholareval}.
Related work has also explored LLMs for time-series forecasting~\citep{gruver2023large,jin2023time,cao2023tempo,ansari2024chronos,das2024decoder,tan2024language,ning2025ts,jiang2025timexl}.
We similarly take a prospective view, but forecast evaluation outcomes for new datasets.


\paragraph{Extracting Structured Records from Literature}
Prior work extracts structured result tuples (e.g., task, dataset, metric) from scientific articles~\citep{singh2019automated, hou2019identification, kardas2020axcell, yang2022telin, bai2023schema, csahinucc2024efficient, park2025can}.
\datasetname\ builds on these tuples: for each experimental record, we synthesize a richer natural-language description that augments the tuple with additional context.

\newcommand{\Yes}{\textcolor{green!55!black}{\ding{51}}} 
\newcommand{\No}{\textcolor{red!70!black}{\ding{55}}}    
\newcommand{\Mix}{\textcolor{violet!70!black}{\ding{55}}}

\begin{table}[t!]
\centering
\small
\setlength{\tabcolsep}{1.2pt}
\begin{tabular}{@{}l c c c c c@{}}
\toprule
& \multicolumn{5}{c}{\textbf{Task Design}} \\
\cmidrule(lr){2-6}
\textbf{Work} & \textbf{Text} & \textbf{Task} & \textbf{Reg.} &  \textbf{No Feat.} & \textbf{No Runs} \\
\midrule
\citet{xia2020predicting}          & \No & \No & \Yes &  \No & \Yes \\ \addlinespace[2pt]
\citet{ye2021finegrained}          & \No & \No & \Yes &  \No & \Yes \\ \addlinespace[2pt]
\citet{ye2023predictable}          & \No & \No & \Yes & \No & \Yes \\ \addlinespace[2pt]
\citet{maiapolo2024tinybenchmarks} & \No & \No & \Yes &  \Yes & \No \\ \addlinespace[2pt]
\citet{pacchiardi2024hundred}     & \No & \No & \Yes &  \No & \No \\ \addlinespace[2pt]
\citet{anugraha2025proxylm}       & \No & \No & \Yes &  \No & \Yes \\ \addlinespace[2pt]
\citet{wen2025predicting}         & \Yes & \No & \No  &  \Yes & \Yes \\ \addlinespace[2pt]
\datasetname~(ours)               & \Yes & \Yes & \Yes &   \Yes & \Yes \\
\bottomrule
\end{tabular}
\caption{Comparison of performance-prediction schemas. \textbf{Text}: uses text descriptions as input; \textbf{Task}: predicts performance on unseen tasks; \textbf{Reg.}: predicts a metric value (regression); \textbf{No Feat.}: requires no manual feature engineering; \textbf{No Runs}: does not require running the model on the benchmark.}
\label{tab:difference}
\end{table}
\section{Task}
\label{sec:task_definition}



\paragraph{Formulation}
We formulate performance prediction from textual descriptions, without access to observed evaluation results, as a regression problem.
Given a self-contained description $x_i$ of a dataset and evaluation configuration, with the dataset name removed,
a model $f_\theta$ predicts a normalized performance score $\hat{y}_i \in [0,100]$ without access to dataset instances or observed evaluation results:
$\hat{y}_i = f_\theta(x_i)$.
The description captures relevant task, dataset, and experimental context in structured yet free-form text.
To compare heterogeneous metrics such as Accuracy and F1, we normalize each paper-reported score $y_i^{\text{paper}}$ to a common 0–100 scale, $y_i=N(y_i^{\text{paper}})$.


\begin{figure*}[t]
  \centering
  \includegraphics[width=1.0\textwidth]{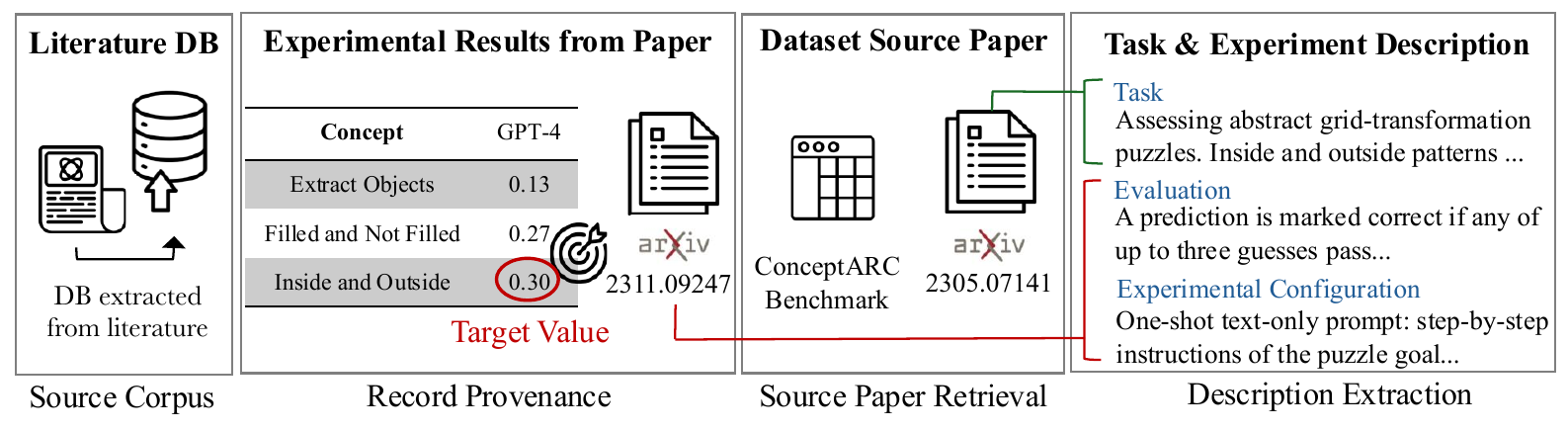}
  \caption{Overview of the \datasetname\ curation pipeline. Each record combines a result paper~\citep{mitchell2023comparing} and a dataset paper~\citep{moskvichev2023conceptarc} to describe the task and evaluation setup. The reported score is normalized to $0$--$100$ as the prediction target.}
  \label{fig:pipeline}
\end{figure*}

\paragraph{Evaluation}
We evaluate the resulting forecasts at two complementary levels: \emph{absolute performance prediction} and \emph{relative ordering}.
For absolute prediction, following prior work~\citep{srinivasan2021predicting, ahuja2022multi, jawahar2024llmpp}, we report mean absolute error (MAE) and Pearson correlation $\mathrm{r}$ between predicted and observed scores.
MAE (lower is better) measures the magnitude of prediction error, while $\mathrm{r}$ (higher is better) measures linear association between predictions and targets.

For relative ordering, we compute pairwise accuracy by comparing predicted scores under matched evaluation settings.
For a pair $(i,j)$, the predicted ordering is correct when
$\operatorname{sign}(\hat{y}_i-\hat{y}_j)=\operatorname{sign}(y_i-y_j)$.
Pairwise accuracy is the fraction of evaluated pairs whose relative ordering is predicted correctly.
Unlike MAE, pairwise accuracy does not require precise score prediction; it measures whether the forecasts preserve which candidate is expected to perform better.
Together, these metrics distinguish the more demanding task of absolute performance prediction from the weaker but potentially useful signal of relative ordering.
 
\section{Data Curation}
\label{sec:data_curation}

To evaluate at scale, we require diverse and realistic description--performance pairs $(x_i,y_i)$. 
Rather than synthesizing experimental configurations and running costly experiments, we curate configurations and reported results directly from the scientific literature. 
As shown in \Cref{fig:pipeline}, our pipeline (i) uses arXiv experimental records, (ii) identifies record provenance and paper types, (iii) retrieves the corresponding sources, (iv) extracts structured descriptions, and (v) performs validation.

\paragraph{Source Corpus}
\label{subsec:source}

We build on LLMEvalDB~\citep{park2025can}, a literature-derived database of LLM evaluation results extracted from experimental tables in arXiv papers, together with their dataset, model, metric, and configurations. 
We use its 2023--2024 entries and 2025 entries, which are dominated by non-finetunable, proprietary models. 
We restrict the target models whose benchmark scores are predicted to GPT-4 and GPT-4o~\citep{achiam2023gpt, hurst2024gpt}.
This choice reduces variation from model and method tuning and better isolates dataset and task effects. 
The 2025 entries also provide a post-cutoff split that helps mitigate leakage.
We treat each reported table cell as an \emph{experimental record} and normalize its score to define $y_i$. 

\paragraph{Record Provenance and Paper Types}
\label{subsec:provenance}

Each record is grounded in two literature sources: the result paper, which reports the score, and the dataset paper, which introduced the dataset. 
These may coincide or differ. 
For example, Paper A introduces a QA dataset and reports accuracy for several models (result paper=dataset paper). 
Alternatively, a different Paper B evaluates a new prompting/evaluation protocol on the same dataset under a revised setup (result paper$\neq$dataset paper)~\citep{park2021faviq, rangapur2024battle}.

\paragraph{Dataset Source Paper Retrieval} 
To compile faithful descriptions for each record, we first collect its sources. LLMEvalDB~\citep{park2025can} provides the result papers. 
When the result paper differs from the dataset paper, we automatically recover the latter by inferring the citation with an LLM, resolving it to an arXiv ID, downloading the PDF, and verifying the dataset name/title. 
We then run a rule-based check to confirm that the PDF explicitly mentions the dataset. 

\paragraph{Extracting Input Description}
\label{subsec:extract_input_feature}

Given the result and dataset papers, an extractor LLM produces a structured description $x_i$ with the dataset name removed.
The description captures the \emph{task} (problem definition and input/label format), \emph{data collection} (provenance, domain, collection protocol), \emph{evaluation} (metric and protocol), \emph{difficulty cues} (e.g., number of classes, long context, reasoning depth), \emph{subset/split}, \emph{prompting strategy} (zero/few-shot, CoT), and other \emph{record-specific settings} (e.g., context window, tool/retrieval use), when reported. See \Cref{appendix:samples} for illustrative examples.

If the retrieved sources do not support the experimental record, the extractor returns \texttt{failed} and we discard the case, providing an additional check against retrieval errors. For valid records, we remove dataset names, numerical results, and summary statistics to produce self-contained, anonymized descriptions.

\paragraph{Data Statistics and Quality Control}
\label{subsec:quality_control}

We construct a bounded yet diverse corpus via stratified sampling over source papers, enforcing caps per paper and dataset to prevent any single source from dominating. 
As shown in \Cref{tab:stats_combined}, this yields 2,290 instances spanning 1,019 dataset papers and 1,022 unique result papers.

Because \datasetname\ is LLM-assisted, we audit 60 descriptions against their source PDFs for anonymization, score leakage, schema coverage, source support, and detail preservation. As shown in \Cref{tab:stats_combined}, 59/60 pass the anonymization check and none leak reported scores, with high schema coverage and source support. Lower detail preservation (3.94/5) suggests that some forecasting-relevant details may be compressed or omitted.

\begin{table}[t!]
\centering
\small
\setlength{\tabcolsep}{3pt}
\begin{tabular}{@{}p{0.65\linewidth}rr@{}}
\toprule
\textbf{Statistic} & \textbf{2023--2024} & \textbf{2025} \\
\midrule
\multicolumn{3}{@{}l}{\textbf{Record counts}}\\
total instances                    & 767  & 1,523 \\
result paper $=$ dataset paper     & 504  & 549  \\
result paper $\neq$ dataset paper  & 263  & 974  \\
\addlinespace[2pt]
\midrule
\multicolumn{3}{@{}l}{\textbf{Paper counts}}\\
unique result paper                & 443  & 579  \\
unique dataset paper               & 471  & 616  \\
result paper $\cup$ dataset paper  & 631  & 956  \\
\midrule
\addlinespace[3pt]
\multicolumn{3}{@{}l}{\textbf{Human validation (60-sample audit)}}\\
Anonymization pass rate             & \multicolumn{2}{r}{59/60 (98.3\%)} \\
Score leakage                       & \multicolumn{2}{r}{0/60} \\
Schema coverage (1--5)              & \multicolumn{2}{r}{4.59} \\
Source support (1--5)               & \multicolumn{2}{r}{4.77} \\
Detail preservation (1--5)          & \multicolumn{2}{r}{3.94} \\
\bottomrule
\end{tabular}
\caption{\datasetname\ statistics by collection period and results of the 60-sample human audit.}
\label{tab:stats_combined}
\end{table} 
\section{Experiments}

We evaluate whether LLMs can forecast benchmark performance from textual descriptions and whether these forecasts provide useful signals for prioritizing candidate evaluations.

\subsection{Experimental Setup}

\paragraph{Baselines} We compare against three families of baselines. 
First, we report the overall \textbf{test-split mean} across all metrics. Because no training or validation set is available, the test-set average serves as a test-set mean oracle baseline.
Second, we evaluate \textbf{text-encoder-based regressors}. We embed each description and train classical models, including \textbf{k-nearest neighbors}~\citep{guo2003knn} and \textbf{XGBoost}~\citep{chen2016xgboost}, on these embeddings. With no dedicated training set, we use 10-fold cross-validation~\citep{fushiki2011estimation}, fitting on nine folds and evaluating on the held-out fold. We use \textbf{E5-Mistral-7B}~\citep{wang2023improving} as the dense embedding encoder. Hyperparameters are tuned on a held-out validation split.

\paragraph{LLMs} We evaluate proprietary and open-source LLMs with and without explicit reasoning. We use \textbf{GPT-5}~\citep{openai2025gpt5}, \textbf{Qwen3-32B}~\citep{yang2025qwen3}, and non-reasoning \textbf{Qwen2.5-32B/72B}~\citep{team2024qwen2}. For GPT-5 and Qwen3, we additionally disable reasoning using \texttt{minimal} reasoning and \texttt{no-thinking}, respectively, to measure its contribution. All models produce a brief rationale followed by a final estimate in \texttt{boxed\{\}}. We use default decoding settings for GPT-5 and the authors' recommended settings for Qwen3.

\subsection{Evaluation Results}
\label{subsection:evaluation_results}

\begin{table}
\centering
\small
\setlength{\tabcolsep}{3pt}
\renewcommand{\arraystretch}{1.12}
\begin{tabular}{@{}lcccc@{}}
\toprule
 & \multicolumn{2}{c}{\textbf{2023--2024}} & \multicolumn{2}{c}{\textbf{2025}} \\
\cmidrule(lr){2-3} \cmidrule(lr){4-5}
\textbf{Model} & \textbf{MAE}~$\downarrow$ & $\mathbf{r}$~$\uparrow$ &
\textbf{MAE}~$\downarrow$ & $\mathbf{r}$~$\uparrow$ \\
\midrule
\multicolumn{5}{c}{\textit{Baselines}} \\
\midrule
Test set average                  & 21.4 & \multicolumn{1}{c}{—} & 23.0 & \multicolumn{1}{c}{—} \\
E5-Mistral + kNN                  & 19.9 & 0.30 & 21.2 & 0.22 \\
E5-Mistral + XGBoost              & 20.3 & 0.30 & 21.4 & 0.20 \\
\midrule
\multicolumn{5}{c}{\textit{LLMs}} \\
\midrule
GPT-5                             & 14.7$^\dagger$ & 0.69 & 14.8$^\dagger$ & 0.62 \\
GPT-5 + \textit{min.-reasoning} & 15.0$^\dagger$ & 0.68 & 15.5$^\dagger$ & 0.52 \\
Qwen3-32B                         & 19.6$^\dagger$ & 0.40 & 19.6$^\dagger$ & 0.41 \\
Qwen3-32B + \textit{no-thinking}  & 22.9 & 0.37 & 22.5 & 0.37 \\
Qwen2.5-32B                       & 23.5 & 0.40 & 21.4 & 0.39 \\
Qwen2.5-72B                       & 22.4 & 0.44 & 19.5$^\dagger$ & 0.43 \\
\bottomrule
\end{tabular}
\caption{Prediction performance on \datasetname\ for 2023–2024 (767) and 2025 (1,523 instances).
$^\dagger$ indicates significantly lower absolute error than the test-set average baseline ($p<0.05$, one-sided binomial sign test~\citep{hollander1999nonparametric}).}
\label{tab:val-results-all}
\end{table}

\paragraph{Performance Forecasting.}

\Cref{tab:val-results-all} summarizes the main prediction results. GPT-5 achieves the lowest MAE on both the 2023--2024 and 2025 splits (14.7 and 14.8), while substantial absolute error remains. Performance is notably similar across the two periods, including the post-cutoff 2025 split, where GPT-5 maintains an MAE of 14.8.

Explicit reasoning improves MAE for both GPT-5 and Qwen3.
The effect is particularly pronounced for Qwen3-32B, whose MAE improves from 22.5 without reasoning to 19.6 with reasoning on the 2025 split. 
However, the performance gap between reasoning and non-reasoning settings is relatively small for GPT-5 models.
In contrast, the text-encoder baselines provide only modest improvements over the global mean, indicating that semantic similarity of benchmark descriptions alone does not fully capture the predictive signal.

\paragraph{Prioritization under Matched Settings.}

We complement absolute-score evaluation by testing whether forecasts correctly rank evaluation configurations under otherwise identical settings.
We construct matched groups sharing the same result paper, dataset, target model, evaluation metric, number of shots, and prompting method, while differing only in the evaluation subset.
Pairwise accuracy measures whether predicted scores correctly order subset pairs within each group.
We additionally evaluate pairs with realized score gaps of at least $\tau$ points; because these gaps are observed only after evaluation, this serves as a post-hoc diagnostic rather than a deployment-time filter.

\begin{figure}[t]
  \centering
  \includegraphics[width=0.95\columnwidth]{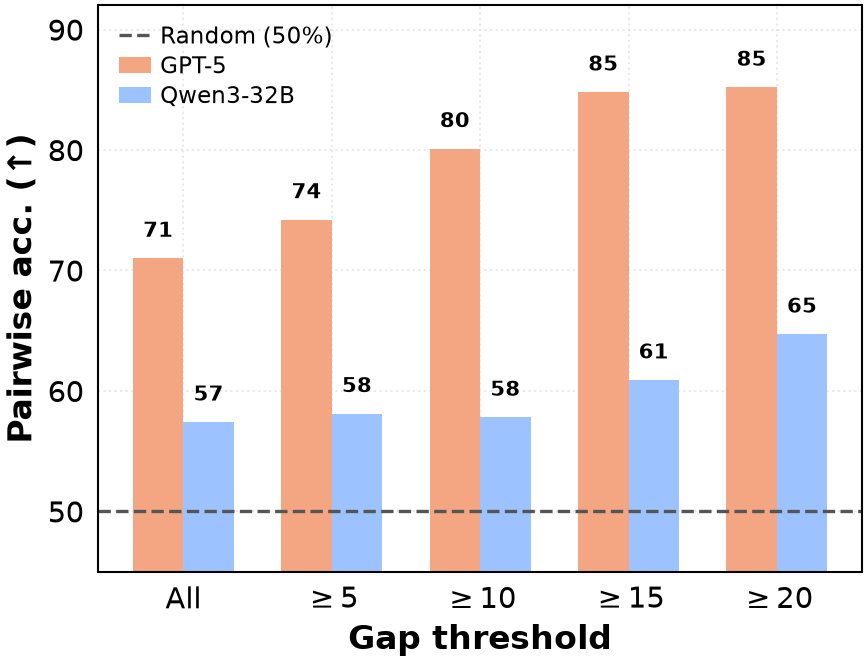}
  \caption{Pairwise accuracy under matched settings across gap thresholds, where each threshold retains only pairs whose realized scores differ by at least that many points. Accuracy measures whether predicted scores correctly preserve the ordering.}
  \label{fig:prioritization}
\end{figure}

As shown in \Cref{fig:prioritization}, GPT-5 substantially exceeds the 50\% random baseline, with pairwise accuracy increasing for larger realized gaps, while Qwen3-32B shows a weaker ordering signal.
A roughly 4--5 point difference in MAE is accompanied by a nearly 20 percentage-point difference in pairwise accuracy, highlighting substantial differences in ranking performance
across predictors.
These results suggest that although accurate absolute performance prediction remains challenging, relative ordering can be more reliably anticipated under matched evaluation settings.
Even when individual score estimates are inaccurate, their relative ordering may still be preserved, highlighting the potential utility for benchmark prioritization.

\begin{table*}[t!]
\centering

\begin{subtable}[t]{0.48\linewidth}
\small
\centering
\caption{By benchmark category.}
\setlength{\tabcolsep}{5pt}
\begin{tabular}{c c cc cc}
\toprule
& & \multicolumn{2}{c}{GPT-5} & \multicolumn{2}{c}{Qwen3} \\
\cmidrule(lr){3-4} \cmidrule(lr){5-6}
\textbf{Category} & \textbf{$n$}
& \textbf{MAE} $\downarrow$ & $\mathbf{r}$ $\uparrow$
& \textbf{MAE} $\downarrow$ & $\mathbf{r}$ $\uparrow$ \\
\midrule
Knowledge
& 970
& \cellcolor{blue!14}14.5 & \cellcolor{blue!8}0.68
& \cellcolor{blue!16}19.3 & \cellcolor{blue!5}0.41 \\

Reasoning
& 1100
& \cellcolor{blue!11}14.8 & \cellcolor{blue!6}0.67
& \cellcolor{blue!14}19.5 & \cellcolor{blue!2}0.37 \\

Math
& 131
& \cellcolor{blue!9}15.0 & \cellcolor{blue!19}0.74
& \cellcolor{blue!2}21.4 & \cellcolor{blue!10}0.45 \\

Coding
& 170
& \cellcolor{blue!8}15.1 & \cellcolor{blue!17}0.73
& \cellcolor{blue!8}20.4 & \cellcolor{blue!20}0.55 \\

Safety
& 63
& \cellcolor{blue!2}15.8 & \cellcolor{blue!19}0.74
& \cellcolor{blue!13}19.6 & \cellcolor{blue!18}0.53 \\

IF
& 382
& \cellcolor{blue!4}15.5 & \cellcolor{blue!5}0.66
& \cellcolor{blue!6}20.7 & \cellcolor{blue!4}0.40 \\

Multilingual
& 60
& \cellcolor{blue!24}12.0 & \cellcolor{blue!24}0.78
& \cellcolor{blue!24}17.9 & \cellcolor{blue!24}0.60 \\
\bottomrule
\end{tabular}
\label{tab:ablation-category}
\end{subtable}
\hspace{0pt plus 0.5em}
%
\begin{subtable}[t]{0.48\linewidth}
\centering
\small
\caption{Peer-reviewed vs.\ non--peer-reviewed.}
\setlength{\tabcolsep}{5pt}
\begin{tabular}{c c c cc}
\toprule
\textbf{Predictor} & \textbf{Test Set} & \textbf{$n$}
& \textbf{MAE} $\downarrow$ & $\mathbf{r}$ $\uparrow$ \\
\midrule
GPT-5
& Peer-reviewed
& 487
& \cellcolor{blue!7}15.4
& \cellcolor{blue!16}0.69 \\

GPT-5
& Not peer-reviewed
& 1036
& \cellcolor{blue!16}14.5
& \cellcolor{blue!12}0.68 \\

Qwen3
& Peer-reviewed
& 487
& \cellcolor{blue!6}20.7
& \cellcolor{blue!8}0.40 \\

Qwen3
& Not peer-reviewed
& 1036
& \cellcolor{blue!16}19.1
& \cellcolor{blue!16}0.42 \\
\bottomrule
\end{tabular}
\label{tab:ablation-peer}
\end{subtable}

\vspace{0.6em}

\begin{subtable}[t]{0.48\linewidth}
\caption{By target model (GPT-4 vs.\ GPT-4o).}
\centering
\small
\begin{tabular}{c c c cc}
\toprule
\textbf{Predictor} & \textbf{Target Model} & \textbf{$n$}
& \textbf{MAE} $\downarrow$ & $\mathbf{r}$ $\uparrow$ \\
\midrule
GPT-5
& GPT-4
& 407
& \cellcolor{blue!16}13.9
& \cellcolor{blue!16}0.73 \\

GPT-5
& GPT-4o
& 1116
& \cellcolor{blue!7}15.2
& \cellcolor{blue!7}0.66 \\

Qwen3
& GPT-4
& 407
& \cellcolor{blue!16}18.1
& \cellcolor{blue!16}0.48 \\

Qwen3
& GPT-4o
& 1116
& \cellcolor{blue!7}20.2
& \cellcolor{blue!7}0.40 \\
\bottomrule
\end{tabular}
\label{tab:ablation-target}
\end{subtable}
\hspace{0pt plus 0.5em}
%
\begin{subtable}[t]{0.48\linewidth}
\caption{Established datasets vs.\ new datasets.}
\centering
\small
\setlength{\tabcolsep}{5pt}
\begin{tabular}{c c c cc}
\toprule
\textbf{Predictor} & \textbf{Test Set} & \textbf{$n$}
& \textbf{MAE} $\downarrow$ & $\mathbf{r}$ $\uparrow$ \\
\midrule
GPT-5
& Established datasets
& 974
& \cellcolor{blue!16}14.7
& \cellcolor{blue!16}0.69 \\

GPT-5
& New datasets
& 549
& \cellcolor{blue!9}15.1
& \cellcolor{blue!8}0.67 \\

Qwen3
& Established datasets
& 974
& \cellcolor{blue!16}19.1
& \cellcolor{blue!16}0.42 \\

Qwen3
& New datasets
& 549
& \cellcolor{blue!7}20.7
& \cellcolor{blue!8}0.40 \\
\bottomrule
\end{tabular}
\label{tab:ablation-setup}
\end{subtable}

\caption{
MAE and Pearson $\mathbf{r}$ across subsets of \datasetname~2025
by benchmark category, publication status, target model, and dataset novelty.
Peer-review status is determined via DBLP;
\textit{new datasets} are introduced in the result paper, and
\textit{established datasets} in prior work.
Darker shading indicates better performance.}
\label{tab:ablation-2x2}
\end{table*}

\subsection{Prediction Heterogeneity Across Subsets}
\label{subsection:evaluation_results_per_metric}

We examine prediction performance across benchmark category, publication status, target model, and established versus new datasets in Table~\ref{tab:ablation-2x2}. 
We define \textit{established datasets} as records where the result paper differs from the dataset paper, and \textit{new datasets} as records where they are the same.

Across categories, GPT-5 and Qwen3-32B show broadly similar trends: multilingual benchmarks have the lowest prediction errors, while math and instruction-following benchmarks tend to have higher errors. 
Both predictors achieve slightly lower MAE on non-peer-reviewed than peer-reviewed benchmarks, although the differences are modest. 
They also achieve lower MAE on GPT-4 than GPT-4o, suggesting that forecasting difficulty varies across target models. 
Finally, established datasets yield slightly lower MAE than new datasets, while correlations remain comparable, suggesting that the predictive signal extends to newly introduced datasets.

\subsection{Pilot Human Study}
\label{subsection:evaluation_results_human}

To contextualize the difficulty of anticipatory performance prediction, we conduct a small pilot study with seven Ph.D. students specializing in NLP and machine learning. Each participant predicts 20 accuracy scores from the same text descriptions provided to the models, yielding 140 predictions.

\begin{figure}[t]
  \centering
  \includegraphics[width=0.89\columnwidth]{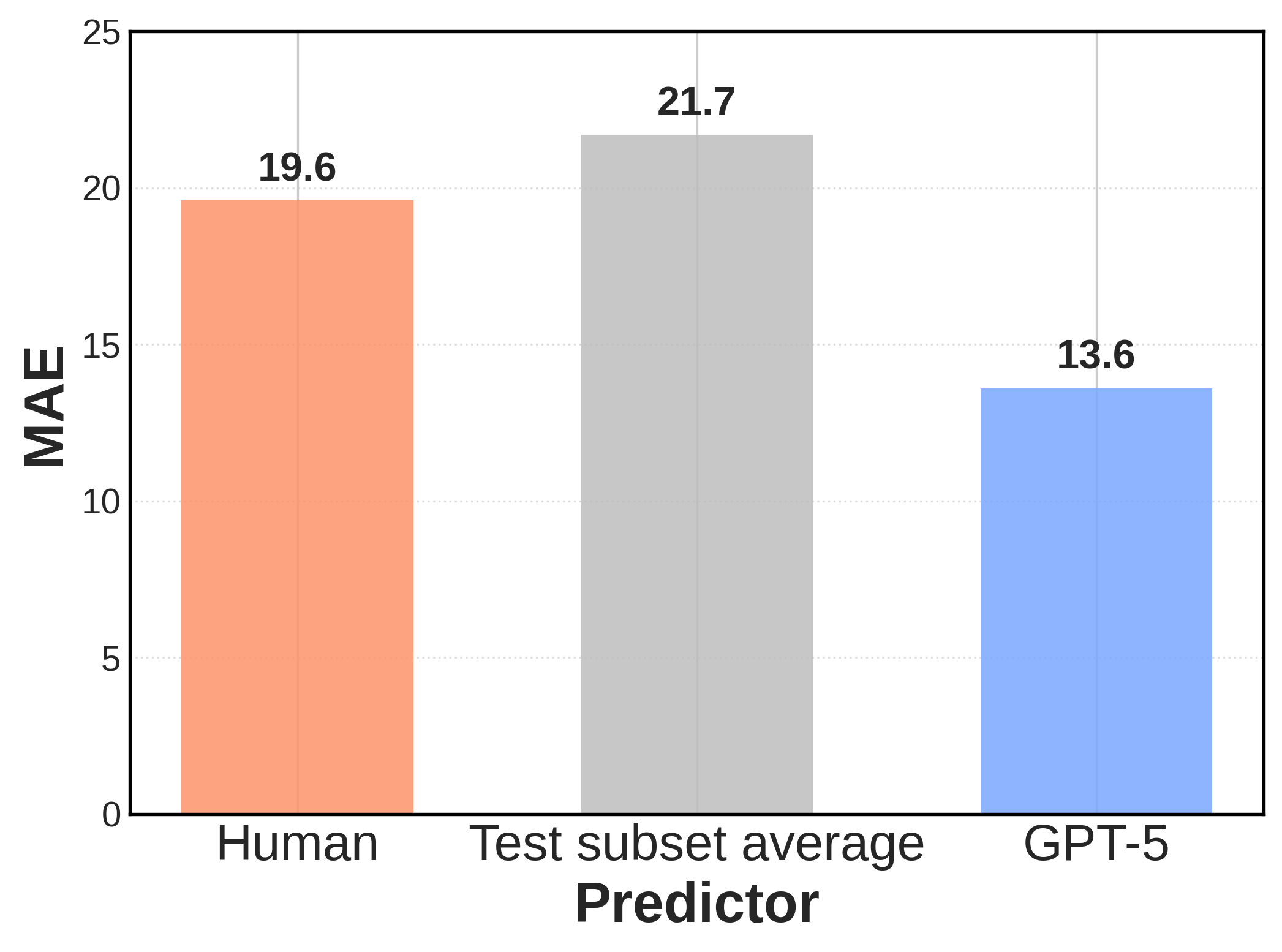}
  \caption{Prediction errors in the pilot human study on the Accuracy subset ($n=140$).}
  \label{fig:human_evaluation}
\end{figure}

As shown in Figure~\ref{fig:human_evaluation}, human predictions achieve $r=0.403$ and an MAE of 19.6, compared with 21.7 for the mean baseline. GPT\text{-}5 achieves a lower MAE on the same subset. Given the small scale of the study, we treat these results as exploratory evidence of task difficulty rather than a general comparison between LLMs and human experts.

\section{Understanding the Prediction Signal}
\label{sec:prediction_analysis}

Beyond aggregate prediction accuracy, we examine whether forecasts respond to controlled task-design changes and characterize recurring failure modes.

\subsection{Sensitivity to Task-Design Cues}
\label{subsec:difficulty_cue}

\begin{table}[t]
\centering
\small
\setlength{\tabcolsep}{3pt}
\renewcommand{\arraystretch}{1.12}
\begin{tabular}{@{}p{0.65\linewidth}r@{}}
\toprule
\textbf{Statistic} & \textbf{Value} \\
\midrule
Counterfactual descriptions   & 250 \\
Median Spearman $\rho$        & $-0.87$ \\
Negative $\rho$               & 50/50 \\
Mean prediction drop          & 15.6 pts \\
\bottomrule
\end{tabular}
\caption{Sensitivity of predicted performance to the number of answer options.}
\label{tab:difficulty_cue}
\end{table}

For 50 multiple-choice or classification examples,
we vary only the number of answer options among
2, 4, 5, 10, and 20, yielding 250 counterfactual
descriptions.
As shown in Table~\ref{tab:difficulty_cue},
predicted performance decreases as the number of
options or classes increases, with a median
Spearman correlation of $-0.87$, negative
correlations for all 50 examples, and a mean
prediction drop of 15.6 points.

These results suggest that the predictor is sensitive to interpretable task-design cues.
However, this experiment measures prediction sensitivity rather than whether the predicted changes match actual changes in target-model performance.

\subsection{Failure Modes of Anticipatory Evaluation}
\label{subsec:error_analysis}


\begin{table}[t]
\centering
\small
\setlength{\tabcolsep}{3pt}
\renewcommand{\arraystretch}{1.12}
\begin{tabular}{@{}p{0.72\linewidth}r@{}}
\toprule
\textbf{Error Category} & \textbf{Share} \\
\midrule
\multicolumn{2}{c}{\textit{Task Characteristics}} \\
\midrule
Ambiguous chance baseline            & 35\% \\
Multi-step compositional reasoning   & 27\% \\
Adversarial or filtered instances    & 13\% \\
Other                                & 25\% \\
\midrule
\multicolumn{2}{c}{\textit{Predictor Failure Modes}} \\
\midrule
Over-anchoring to chance performance & 34\% \\
Underestimating compounding errors   & 32\% \\
Overgeneralizing capability priors   & 22\% \\
Other                                & 12\% \\
\bottomrule
\end{tabular}

\caption{Error taxonomy for the 100 largest Accuracy prediction errors. Each case is categorized by both task characteristic and predictor failure mode.}

\label{tab:error_taxonomy}

\end{table}

We manually examine the 100 largest Accuracy prediction errors, assigning each case a task characteristic and a predictor-side failure mode. 
As shown in \Cref{tab:error_taxonomy}, errors frequently occur in tasks with ambiguous chance baselines, where the label space does not imply a straightforward random-guessing accuracy, and in multi-step compositional tasks, where success requires several dependent reasoning steps. 
Common predictor failures include over-anchoring to chance performance, underestimating compounding errors across reasoning steps, and assuming that a model’s general capabilities transfer equally well across different tasks. 
These patterns highlight recurring limitations in precise performance estimation.


\begin{figure*}[t!]
  \centering
  \includegraphics[width=0.9\textwidth]{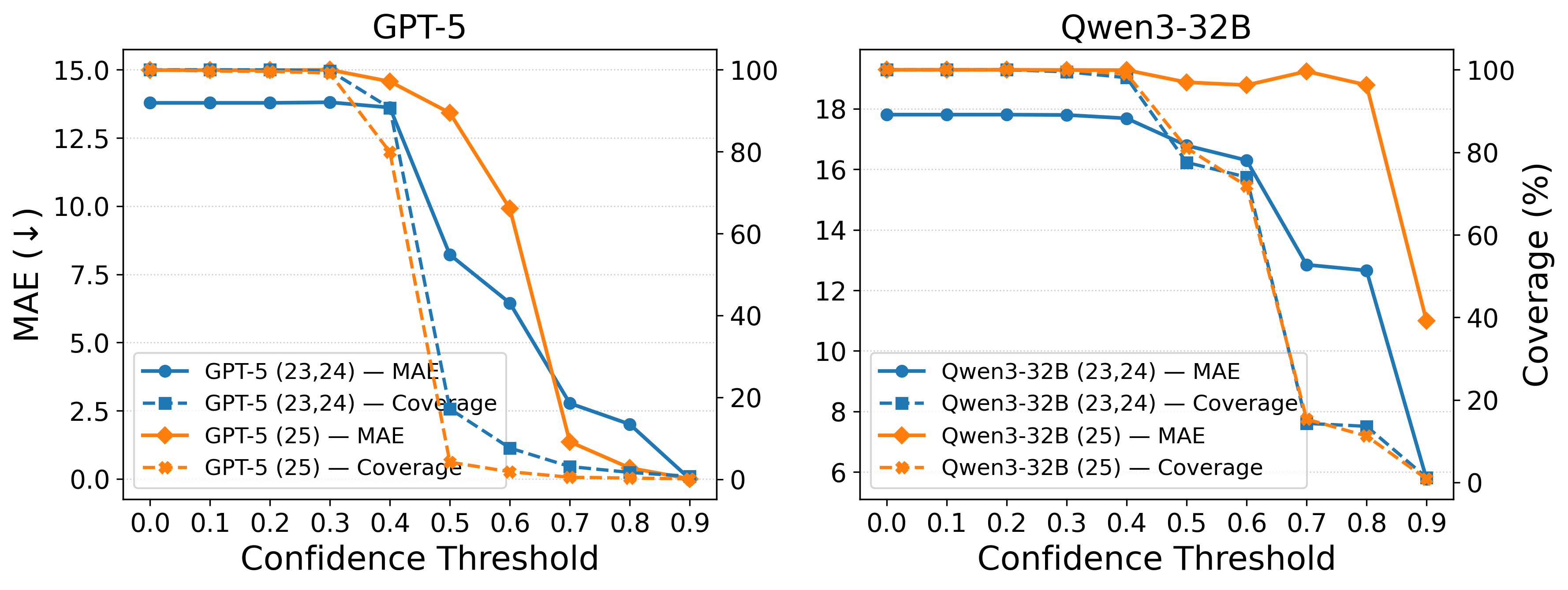}
  \caption{MAE (left axis) and coverage (right axis) as a function of confidence thresholds (scores $\geq$ band lower bound) for GPT-5 and Qwen3-32B across different year splits (2023-2024 vs 2025). Higher thresholds retain fewer examples (lower coverage) and generally yield lower MAE. Confidence values are derived from a 10-bin categorical output elicited from the model.}
  \label{fig:calibration_results}
\end{figure*}

\section{Retrieval Analysis}

We assess the contribution of retrieval using tool-call counts and query token diversity (tokenized with GPT-4o~\citep{hurst2024gpt}). All models use a custom retriever over the arXiv API,\footnote{\url{https://info.arxiv.org/help/api/index.html}} with each record's source papers excluded by identifier. We initially tested GPT-5's built-in web search, but an audit found that roughly 70\% of evaluation source papers became identifiable, motivating the use of the controlled retriever instead. Each query returns at most two documents, with up to four retrievals under ReAct-style prompting~\citep{yao2023react}. We compare retrieval-enabled and description-only settings to measure the contribution of literature grounding.


As shown in Table~\ref{tab:search_token_stats}, retrieval
improves GPT-5 but not Qwen3.
GPT-5 makes more search calls with more diverse
queries, while Qwen3 uses retrieval more sparsely.
Retrieved documents are typically only moderately
relevant on a 5-point LLM-rated scale, which may
explain the limited gains from retrieval.
This contrast indicates that access to external information alone does not guarantee better forecasts; its usefulness may depend on how effectively a predictor incorporates relevant evidence. Overall, retrieval provides a modest, model-dependent benefit rather than being the primary source of predictive performance.

\begin{table}[t!]
\centering
\small
\setlength{\tabcolsep}{3pt}
\begin{tabular}{@{}p{0.55\linewidth}rr@{}}
\toprule
& GPT-5 & Qwen3-32B \\
\midrule
\multicolumn{3}{@{}l}{\textit{Prediction performance} (\textbf{MAE} / $\mathbf{r}$)}\\
No \textit{Search}      & 14.7 / 0.69 & 19.6 / 0.40 \\
With \textit{Search}  & 14.0 / 0.73 & 20.8 / 0.31 \\
\midrule
\multicolumn{3}{@{}l}{\textit{Search-tool usage \& token diversity}}\\
Avg.\ search calls         & 3.06 & 0.89 \\
Std.\ search calls         & 0.87 & 0.32 \\
Unique tokens (all)        & 2,140 & 794 \\
Unique tokens (random)     & 1,321 & \textemdash \\
\bottomrule
\end{tabular}
\caption{Search-tool usage, token diversity, and prediction performance for GPT-5 and Qwen3-32B. Token counts are measured with the GPT\text{-}4o tokenizer. (all) aggregates all formulated queries; (random) uses one random query per instance.} 
\label{tab:search_token_stats}
\end{table}

\section{Confidence and Prediction Error}



Some experimental outcomes may be inherently difficult
to predict, raising the question of whether models can
reliably estimate their own prediction accuracy.
To examine this, we elicit a scalar, self-reported
confidence for each prediction (following
\citet{yoon2025reasoning}) and examine prediction error
among items with confidence $\geq \tau$.
For each $\tau$, we report the MAE on the retained
subset together with its coverage, which gives
a view of the error--coverage trade-off.
We report confidence curves for the accuracy subset.

Figure~\ref{fig:calibration_results} shows results for GPT-5
and Qwen3-32B on the accuracy subset, separately
for the 2023--2024 and 2025 splits.
As the confidence threshold increases, coverage
declines and MAE tends to decrease, indicating
that higher reported confidence is, on average,
associated with more accurate forecasts.
MAE decreases more sharply on the 2023--2024 split
as the confidence threshold increases, possibly
reflecting greater training-time exposure to
similar examples from this period.

These results suggest that self-reported confidence may help identify subsets of predictions with lower observed error.
However, this comes at the cost of reduced coverage,
and the observed relationship does not establish
that confidence scores are well calibrated. 
\section{Streaming Prediction}
\label{sec:streaming_prediction}


\begin{figure}[t]
  \centering
  \includegraphics[width=0.89\columnwidth]{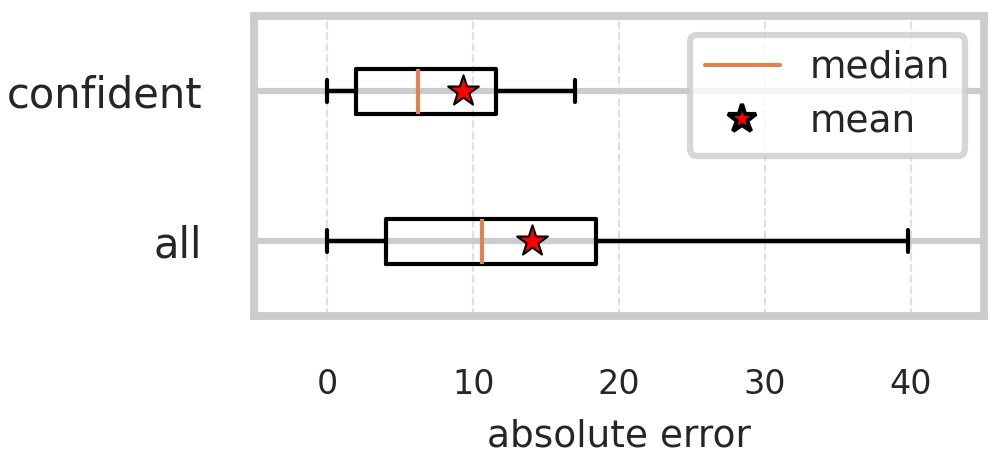}
  \caption{Streaming prediction errors for GPT\text{-}5 + Search. \textit{Confident} denotes predictions with over 50\% confidence~\citep{yoon2025reasoning}.}
  \label{fig:streaming_box_plot}
\end{figure}

Although the post-cutoff split reduces leakage risk, indirect exposure to reported results cannot be fully excluded. 
We therefore evaluate a \emph{streaming prediction} setting, forecasting newly reported arXiv results immediately after release and before third-party search indexing.

\paragraph{Setup}
We collect newly released papers from arXiv’s official interfaces for \texttt{cs.CL}, \texttt{cs.LG}, and \texttt{cs.AI} and process them through the same extraction pipeline (see \Cref{app:arxiv-compliance}). 
We restrict records to our seven metric groups and evaluate GPT-family models. Built-in web search is permitted, but predictions are excluded if the target paper is retrieved, as detected by arXiv ID and content matching. 
During the prospective collection window, we obtain 79 previously unseen experimental records from 52 papers and issue predictions before third-party search indexing. 
Exact collection dates are reported in \Cref{app:arxiv-compliance}.

\begin{table}[!t]
\centering
\small
\setlength{\tabcolsep}{5pt}
\renewcommand{\arraystretch}{1.12}
\begin{tabular}{@{}lcc@{}}
\toprule
\textbf{Model} & \textbf{MAE}~$\downarrow$ & $\mathbf{r}$~$\uparrow$ \\
\midrule
Test set average      & 25.5 & --   \\
GPT\text{-}5          & 15.0 & 0.74 \\
GPT\text{-}5 + Search & 14.1 & 0.78 \\
\bottomrule
\end{tabular}
\caption{Prediction performance in the streaming setting on 79 experimental records.}
\label{tab:streaming_prediction}
\end{table}

\paragraph{Results}
As shown in \Cref{tab:streaming_prediction}, GPT\text{-}5 + Search achieves an MAE of 14.1 and $r=0.78$ on newly released records, comparable to its performance on \datasetname. 
The high-confidence subset ($>$50\% confidence; 23 samples) has a lower average absolute error of 9.3 (\Cref{fig:streaming_box_plot}). 
In an additional self-referential test conducted before this paper was searchable, GPT\text{-}5 + Search predicted an MAE of 16.0 for the task in this paper, compared with the realized 14.7 in \Cref{tab:val-results-all}. Together, these prospective evaluations provide additional evidence that the observed forecasting signal is not solely attributable to exposure to previously reported results.
 
\section{Discussion and Conclusion}
\label{sec:conclusion}

We introduce \emph{anticipatory evaluation} for predicting benchmark performance before experiments, together with
\datasetname.
Our results show that while accurate absolute score
prediction remains challenging, LLMs can provide useful
signals about relative benchmark difficulty.
These signals persist across post-cutoff and streaming evaluations, suggesting they are not solely attributable to exposure to reported results.
Further analyses reveal sensitivity to cues
and recurring failure modes.


These findings point to a distinction between forecasting
exact performance and identifying relative differences
across candidate evaluations.
Rather than relying on precise score estimates,
researchers may use coarse forecasts to inform which
benchmarks warrant further empirical investigation.
Such forecasts may be particularly useful when
comparing evaluation variants under similar settings,
where relative difficulty matters more than exact scores.
Future work could evaluate their practical utility
in guiding benchmark development.

\newpage

\section*{Limitations}

\paragraph{Planning-oriented scope.}
Our method is intended as a lightweight planning tool for estimating task difficulty, prioritizing candidate experiments, and designing pilot studies before substantial evaluation costs are incurred. Its predictions should be interpreted as coarse decision support rather than substitutes for empirical evaluation. We also do not conduct a prospective user study of benchmark design; future work should evaluate whether anticipatory predictions improve real design decisions or reduce evaluation cost.

\paragraph{Focus on the dataset axis.}
Our study mainly focuses on the dataset axis, with target models limited to GPT-4 and GPT-4o. Generalization to substantially different model families, scales, or training paradigms, including smaller and open-source models, therefore remains unverified.

\paragraph{Dependence on reconstructed descriptions.}
\datasetname\ uses descriptions reconstructed from completed papers, which are not equivalent to actual early-stage proposals where relevant details may be unspecified. Forecast quality depends on the completeness and specificity of these descriptions, and our human audit indicates that some forecasting-relevant details may be compressed or omitted during reconstruction.

\paragraph{Correlational rather than causal signals.}
The method does not establish that the signals driving predictions are causally related to downstream performance. Predictions may rely on correlational regularities in benchmark descriptions, reporting conventions, or recurring task structures, limiting their use to coarse prioritization rather than precise empirical conclusions.

\paragraph{Residual leakage risk.}
Despite post-cutoff evaluation, streaming-style settings, and source-grounded auditing, indirect leakage cannot be fully ruled out. Benchmarks and models may still be connected to training-time exposure through repeated descriptions, summaries, or widely circulated benchmark narratives.

\section*{AI Usage}

LLM-based tools were used to assist with grammar checking,
wording refinement, and figure-generation code during
manuscript preparation. All AI-assisted content was
reviewed and verified by the authors.

\bibliography{acl_latex}

\newpage
\appendix
\onecolumn

\section{Details of Data Curation}
\label{appendix:data_curation}

This section provides data-curation details not covered in \Cref{sec:data_curation}.

\paragraph{LLMs}
For curation we used GPT-5-mini~\citep{openai2025gpt5}. 
The model extracted schema-aligned descriptions from source papers and handled dataset-source retrieval annotations. 
These choices were based on a pilot run and seed-result quality.

\paragraph{Dataset-source arXiv annotation}
Given the result paper PDF, we prompted GPT-5-mini to infer the title of the dataset-source paper, queried the arXiv API to resolve an identifier, and downloaded the PDF. 
We improve precision by automatically checking content (e.g., verifying that the dataset name appears in the retrieved paper).

\paragraph{Description extraction}
We excluded appendix tables during extraction. 
To disambiguate records, we supplied the pre-normalization metric value as context. 
For efficiency, all PDFs were truncated to the first 12 pages during retrieval and extraction.

\paragraph{Additional filtering}
All records involve GPT-4~\citep{achiam2023gpt} or GPT-4o~\citep{hurst2024gpt}, fixing the model/method axis to focus on dataset/task variation. 
We excluded multimodal domains, correlation-only metrics (not on a 0–100 scale), and overall benchmark aggregates (e.g., MMLU) via heuristic filters.
\section{Annotation Guidelines}
\label{appendix:annotation_guidelines}

\subsection{Description Quality}

This section details the human annotation protocol used to evaluate the quality of redacted, schema-aligned descriptions against their source PDFs. 
Annotators were instructed to verify that each description is anonymized, complete where possible, and strictly grounded in the provided sources. 

\paragraph{Annotation Task}
Annotators were given:
\begin{itemize}[leftmargin=1.8em]
    \item The extracted input description for a given experimental record.
    \item The corresponding result and dataset paper PDFs.
\end{itemize}
The task was to check whether the description faithfully reflects information from the PDFs, is anonymized, and adheres to the fixed schema.

\subsection{Evaluation Criteria}
Three dimensions of quality were assessed:

\paragraph{Anonymization (Binary).}
Descriptions must not include dataset or subset names, and must exclude performance numbers drawn from the main paper. 
Annotators assign either \textbf{Accept} (score: 1.0) or \textbf{Reject} (score: 0.0).  

\paragraph{Schema coverage (1--5).}
Checks whether all schema fields are filled appropriately when information is available in the PDFs, and ``None'' is used only when the information is truly absent.  
\begin{itemize}[leftmargin=1.5em]
    \item \textbf{Score 1}: Ignores the schema; most fields missing or collapsed.  
    \item \textbf{Score 2}: Very few fields filled; many omissions; ``None'' often misused.  
    \item \textbf{Score 3}: About half of fields filled; some omissions or misplacements.  
    \item \textbf{Score 4}: Nearly all fields correctly covered; rare omissions or minor misplacements.  
    \item \textbf{Score 5}: Complete and consistent coverage; ``None'' only when truly absent; perfectly aligned with schema.  
\end{itemize}

\paragraph{Factual grounding (1--5).}
Ensures every statement in the description is directly supported by the PDFs, without speculation or outside knowledge.  
\begin{itemize}[leftmargin=1.5em]
    \item \textbf{Score 1}: Mostly unsupported or invented; contradictions with PDFs.  
    \item \textbf{Score 2}: Frequent unsupported claims; weak sourcing.  
    \item \textbf{Score 3}: Generally grounded, but several unsupported claims or mild over-interpretation.  
    \item \textbf{Score 4}: Fully grounded with only minor overstatements; no contradictions.  
    \item \textbf{Score 5}: Entirely faithful; precise paraphrases of the PDFs; every fact traceable.  
\end{itemize}

\subsection{Human Performance Prediction}

This subsection specifies the human annotation protocol for predicting task performance from redacted, schema-aligned descriptions. Annotators produce a single accuracy estimate and a brief rationale \emph{without consulting external sources}, enabling an assessment of human forecasting difficulty and behavior under text-only constraints.

\paragraph{Annotation Task}
Annotators are given:
\begin{itemize}[leftmargin=1.8em]
    \item A redacted description of an experimental setup (model, dataset/task, evaluation protocol, and salient configuration details).
    \item A fixed-time budget per item and an interactive prompt that records their identity (first name) and assigns a shard automatically.
\end{itemize}
For each item, annotators must output:
\begin{itemize}[leftmargin=1.8em]
    \item \textbf{Predicted accuracy} (integer in 1--100).
    \item \textbf{Rationale} (1--3 sentences) briefly justifying the estimate.
\end{itemize}

\paragraph{Instructions and Constraints}
\begin{itemize}[leftmargin=1.8em]
    \item \textbf{No external search.} Rely solely on the provided description and personal prior knowledge; do not use web search, papers, or tools.
    \item \textbf{Timing.} Aim to complete each prediction within 2--3 minutes.
    \item \textbf{Batch size.} Each annotator completes 20 samples per session.
\end{itemize}
\section{Supplementary Experimental Results}

\begin{figure}[t]
  \centering
  \includegraphics[width=1.0\columnwidth]{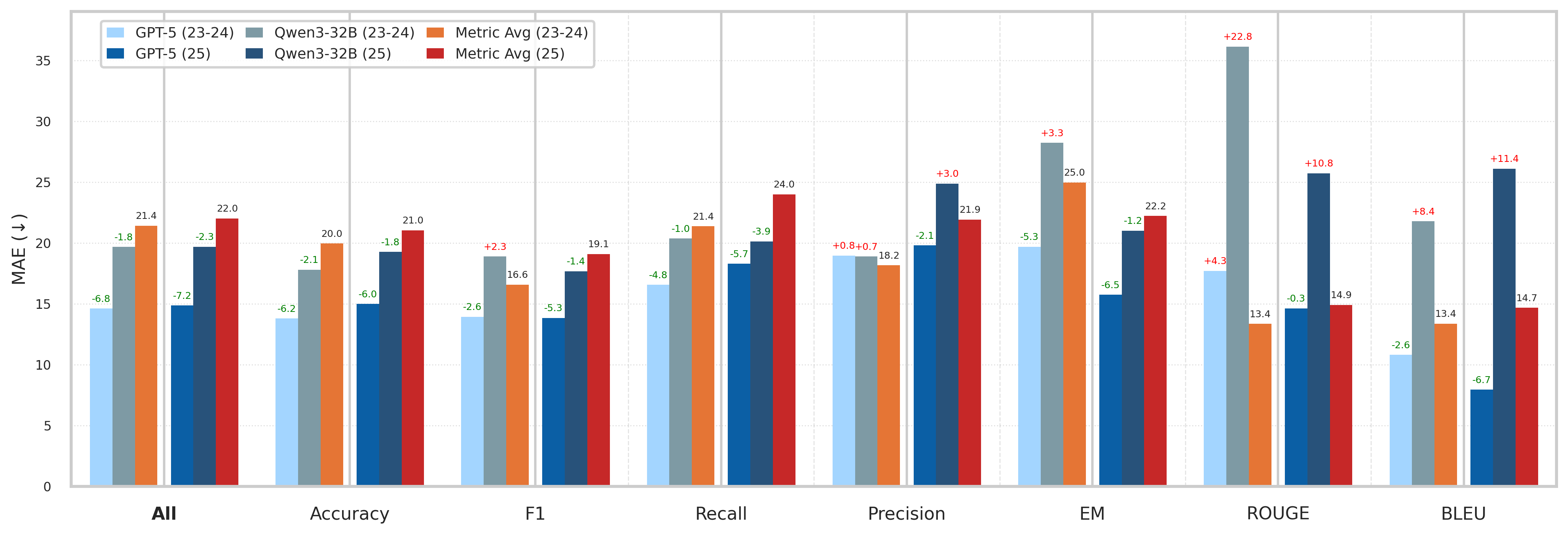}
  \caption{Per-metric MAE (↓) for all baselines and LLMs, grouped by evaluation metric (e.g., Accuracy, Exact Match (EM), Recall, F1) and computed separately on the 2023–2024 and 2025 data subsets. The red and orange bars show the per-metric averages for each year split, with their absolute MAE values labeled, and the remaining bars indicate how much each model deviates from the corresponding yearly average.
  }
  \label{fig:evaluaion_results_per_metric_full}
\end{figure}

\paragraph{Per Metric MAE for all the baselines \& LLMs} We report per-metric MAE for all baselines and LLMs, along with a test-calibrated per-metric mean baseline across seven metric groups in \Cref{fig:evaluaion_results_per_metric_full}.

\paragraph{Additional results with newer embedding backbones.}
We ran additional experiments on the \textsc{\datasetname-2025} dataset using more recent embedding models that are competitive or state-of-the-art on the MTEB leaderboard, including \textsc{Qwen3-8B-Embed}~\citep{zhang2025qwen3} and \textsc{KaLM-Embedding-Gemma3-12B-2511}~\citep{hu2025kalmembedding}, combined with both KNN and XGBoost regressors. 
The results are summarized in Table~\ref{tab:precog_2025_new_backbones}. 
As the table shows, newer and stronger embedding models yield modest improvements in MAE and correlation, but the gains are relatively small. 

\begin{table}[t]
\centering
\small
\begin{tabular}{lcc}
\toprule
Model & MAE & Pearson Correlation \\
\midrule
E5-Mistral-7B + KNN                      & 21.2 & 0.22 \\
E5-Mistral-7B + XGBoost                  & 21.4 & 0.20 \\
Qwen3-8B-Embed + KNN                     & 20.8 & 0.27 \\
Qwen3-8B-Embed + XGBoost                 & 21.6 & 0.21 \\
KaLM-Embedding-Gemma3-12B-2511 + KNN     & 20.8 & 0.26 \\
KaLM-Embedding-Gemma3-12B-2511 + XGBoost & 21.5 & 0.21 \\
\bottomrule
\end{tabular}
\caption{Results on the \textsc{\datasetname-2025} dataset with different embedding backbones and regressors. Newer, stronger embedding models provide modest gains, but improvements are relatively small compared to the overall bottleneck from compressing each model--task pair into a fixed-size vector.}

\label{tab:precog_2025_new_backbones}
\end{table}

\section{Illustrative Case Study Example}

\subsection{Puzzle QA case study (on-air vs.\ off-air puzzles).}

We consider a benchmark of verbal reasoning puzzles drawn from a long-running radio show~\cite{wu2025phd}. 
The original authors distinguish two settings: (i) on-air puzzles, which are short, pattern-based word questions intended to be solved live in a few seconds by callers during the broadcast, and (ii) off-air weekly challenges, which are substantially harder puzzles that listeners may think about for days before submitting an answer. 
In the released benchmark, GPT-4o achieves around 65\% accuracy on the easier on-air subset, but only about 6\% on the more challenging off-air subset.

\begin{table}[t]
\small
\centering
\begin{tabular}{lcc}
\toprule
Subset & Gold / Pred.\ acc.\ (\%) & Abs.\ error (pp) \\
\midrule
On-air  & 65 / 68  & 3 \\
Off-air &  6 / 14  & 8 \\
Overall & \textemdash & 5.5 (MAE) \\
\bottomrule
\end{tabular}
\caption{Gold and predicted GPT-4o accuracies on the on-air and off-air puzzle subsets, along with absolute error in percentage points (pp).}
\label{tab:puzzle_on_off_air}
\end{table}

To simulate how our system could assist benchmark design \emph{before} either subset is fully constructed, we ignore these empirical numbers and assume a designer only has two candidate formulations in mind. The designer writes two short, detailed descriptions: an on-air description emphasizing that puzzles are designed for live radio, solvable in seconds, with strongly structured patterns and relatively shallow wordplay; and an off-air description emphasizing that puzzles are intended as week-long challenges, often involving multi-step verbal reasoning, broader general knowledge, and a much larger search space over possible candidate answers. 
These natural-language descriptions, together with the target model name, are provided as the only inputs to our predictor; crucially, no performance statistics from the original benchmark are revealed.

Under this purely description-based setup, our system predicts 68\% accuracy for GPT-4o on the on-air variant and 14\% on the off-air variant as in Table~\ref{tab:puzzle_on_off_air}. When compared to the eventual empirical scores (65\% and 6\%), the corresponding absolute errors are 3 and 8 percentage points, respectively, yielding a mean absolute error of 5.5 percentage points. The model correctly predicts a large drop in performance going from on-air to off-air.

From a benchmark designer's perspective, this ex-ante signal is already actionable. Suppose the designer's goal is to avoid both saturation (a task that is too easy for current models) and degeneracy (a task on which models essentially always fail). The predicted 68\% versus 14\% suggests that the on-air formulation is likely to be quickly saturated by strong models, whereas the off-air formulation is unlikely to be impossible (for example, 0\% accuracy), and instead lies in a more interesting, mid-range difficulty regime. In a hypothetical setting where neither dataset yet exists, the designer could draft natural-language specifications for an ``easy/live'' and a ``hard/take-home'' variant, query our system once to obtain rough performance predictions on each, and then use these predictions to prioritize constructing and cleaning the more promising hard variant (off-air) first. Only that variant would need to be validated initially with empirical model runs, reducing the need to fully curate both candidate datasets and run pilot experiments on each. While our predictions still require downstream validation, they provide a quantitative shortcut that can guide early benchmark design decisions, turning vague intuitions (``off-air seems harder'') into concrete, model-specific expectations.

\begin{figure}[t]
  \centering
  \includegraphics[width=0.6\columnwidth]{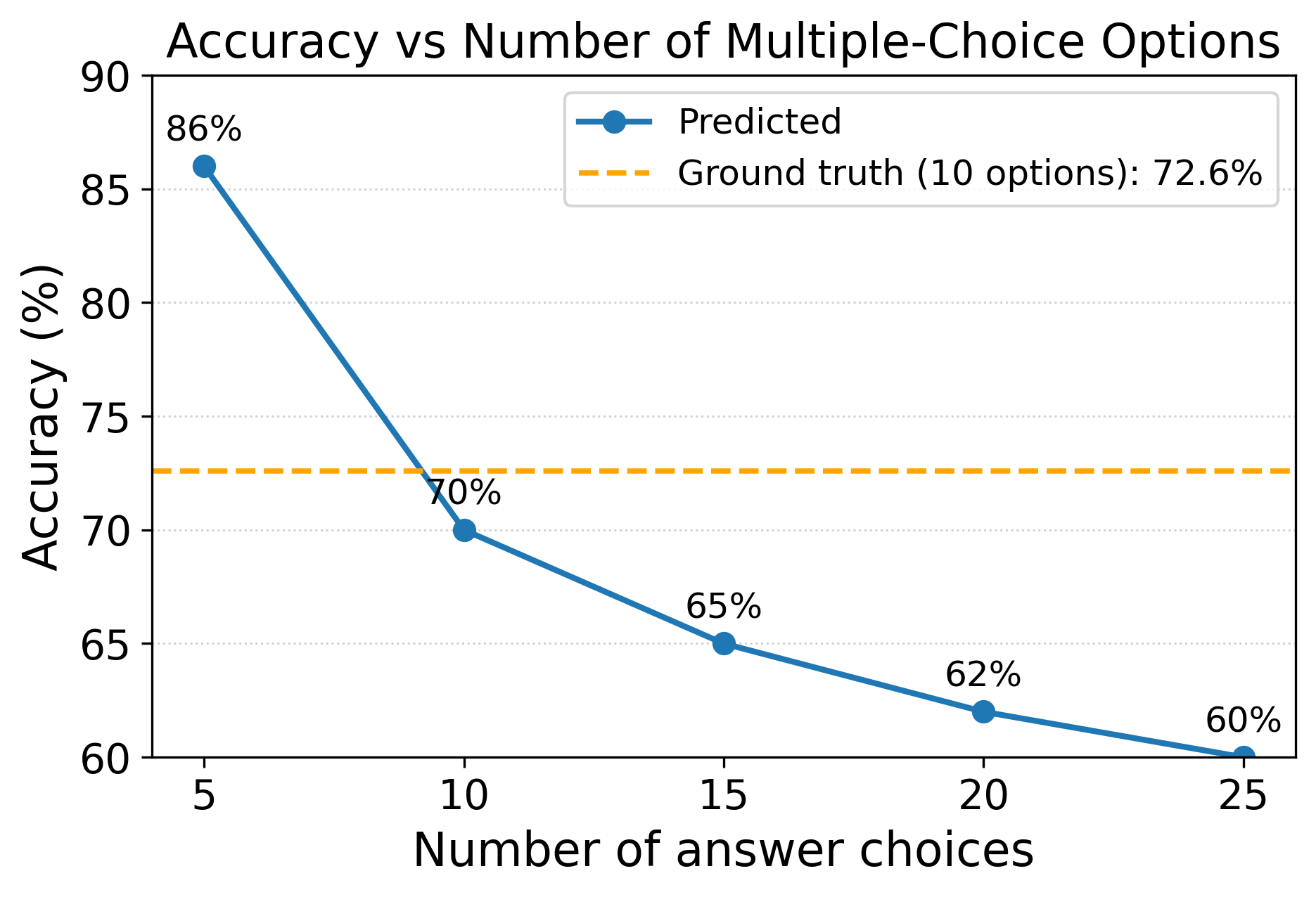}
  \caption{Predicted GPT\textendash4o accuracy as a function of the number of multiple-choice options for a multitask reasoning benchmark. The solid line shows our description-based predictions for 5, 10, 15, 20, and 25 options, and the dashed horizontal line marks the reported ground-truth accuracy of 72.6\% for the 10-option configuration used in practice.
  }
  \label{fig:mmlu_options}
\end{figure}

\vspace{-.1cm}

\subsection{MMLU-Pro: Choosing the number of answer options}

To illustrate how description-based prediction can support design-time decisions, we consider a multitask multiple-choice benchmark in the style of MMLU-Pro~\citep{wang2024mmlu}, and focus on a single design configuration: the number of answer options per question. 
We imagine a benchmark designer who has not yet built the dataset and is deciding between 5, 10, 15, 20, or 25 options. 
For each configuration, the designer writes a natural language description (same questions and domains, different number of options) and queries our predictor, which estimates GPT\textendash4o~\citep{hurst2024gpt} accuracy from text alone, without access to any items.

Figure~\ref{fig:mmlu_options} summarize the results. Using only these textual descriptions, our predictor (GPT\textendash5) estimates GPT\textendash4o accuracies of 86\%, 70\%, 65\%, 62\%, and 60\% for 5, 10, 15, 20, and 25 options, respectively, while the reported ground truth accuracy at 10 options is 72.6\% from the original paper~\citep{wang2024mmlu}. 
The predictions are monotone in the expected direction and reasonably calibrated at the configuration used in practice.

This example shows how a single batch of zero-example predictions can already guide a practical choice. Before collecting any data, the designer can (i) rule out clearly saturated settings such as 5 options, (ii) avoid very difficult settings such as 20--25 options that risk making the benchmark unusable for current models, and (iii) focus annotation and compute on intermediate configurations around 10--15 options where performance is expected to fall in a useful range. In this sense, our method provides a coarse but actionable quantitative signal for prioritization and resource allocation at design time.

\newpage

\subsection{on-air setting description}

\definecolor{titleblue}{RGB}{38,116,186}
\definecolor{frameblue}{RGB}{38,116,186}
\definecolor{backblue}{RGB}{240,247,255}

\tcbset{
  mybox/.style={
    enhanced,
    breakable,
    colback=backblue,
    colframe=frameblue,
    boxrule=0.8pt,
    arc=3mm,
    left=10pt,right=10pt,top=10pt,bottom=10pt
  },
  mytitle/.style={
    fontupper=\bfseries\large\color{white},
    colback=titleblue,
    colframe=titleblue,
    boxrule=0pt,
    arc=2mm,
    left=8pt,right=8pt,top=6pt,bottom=6pt
  }
}

\lstdefinestyle{tightmono}{
  basicstyle=\ttfamily\small,
  columns=fullflexible,
  keepspaces=true,
  showstringspaces=false,
  breaklines=true,
  frame=single,
  framerule=0.3pt,
  rulecolor=\color{black!35},
  backgroundcolor=\color{white},
  tabsize=2
}

\setlist[itemize]{topsep=3pt,itemsep=2pt,parsep=0pt,leftmargin=1em,labelsep=0.5em}
\setlength{\parskip}{4pt}
\setlength{\parindent}{0pt}

\begin{tcolorbox}[mybox]

\textbf{Model:} GPT-4o

\begin{itemize}
  \item \textbf{Task:}
    \begin{itemize}
      \item Open-ended answer generation for short word puzzles presented in a structured format.
      \item Input: a shared natural-language description that specifies a wordplay pattern or rule (e.g., a relationship between two words), plus one concrete question that instantiates this pattern.
      \item Answer space: unconstrained natural-language strings, typically a single word or very short phrase (for example, a word that becomes a synonym of another after a letter rearrangement).
      \item Evaluation target: determine whether the model’s generated text contains the correct target word or phrase that satisfies the described constraint for that question.
    \end{itemize}

  \item \textbf{Data collection:}
    \begin{itemize}
      \item Derived from a recurring live puzzle segment on a radio program (on-air).
      \item In each segment, a host first reads a brief description that defines a wordplay pattern, then poses a sequence of individual questions, each consisting of concrete words or phrases that fit the pattern; callers attempt to solve these in real time during the broadcast.
      \item For each description, multiple question–answer pairs are associated with the same underlying pattern, yielding clusters of structurally similar items.
      \item These on-air questions and their solutions were previously collected and standardized into a benchmark by earlier work, and are reused here to evaluate modern models; the present setup relies on that standardized corpus rather than reconstructing the scraping and transcription pipeline.
    \end{itemize}

  \item \textbf{Evaluation:}
    \begin{itemize}
      \item Metric: accuracy over individual question–answer pairs.
      \item Each on-air question is treated as one evaluation item; the model is prompted with the description and the corresponding concrete question, and produces a free-form completion.
      \item A completion is counted as correct if it contains the gold solution string for that question, ignoring superficial differences such as capitalization and punctuation.
      \item No partial credit is awarded: items are marked simply correct or incorrect, and the reported score is the mean accuracy over all on-air questions in this subset.
    \end{itemize}

  \item \textbf{Difficulty characteristics:}
    \begin{itemize}
      \item Puzzles are explicitly designed to be solvable within a few seconds by human callers during a live broadcast, making them easier than the show’s separate, week-long take-home challenges.
      \item The shared description typically makes the pattern very explicit (for example, specifying exactly how two given words must be related), and the associated questions closely follow this pattern, providing strong guidance about the required transformation or relationship.
      \item Most items involve letter-level or word-level constraints (such as rearranging letters to obtain a synonym), sometimes combined with simple semantic relations (e.g., basic synonymy), rather than deep multi-step reasoning.
      \item Because the same description applies to multiple questions, the underlying reasoning structure is repeated within small clusters; once the model infers the pattern from the description, each additional question in that cluster is another instance of the same structure, reducing overall difficulty relative to more heterogeneous, one-off puzzles.
    \end{itemize}

  \item \textbf{Subset:}
    \begin{itemize}
      \item This record corresponds specifically to the on-air subset of puzzles from the radio program.
      \item Every item in this subset is derived from questions intended to be answered live during the broadcast under tight time constraints.
      \item The subset excludes the separate weekly take-home challenges and does not apply further filtering by puzzle type or knowledge category beyond what was done in the original benchmark construction.
    \end{itemize}
\end{itemize}

\begin{itemize}
  \item \textbf{Prompting strategy:}
    \begin{itemize}
      \item Zero-shot generation on the on-air puzzles.
      \item For each question–answer pair, the prompt consists of the textual description of the pattern together with the specific instantiation (the question containing the concrete words or phrases).
      \item No in-context examples from other puzzles, no few-shot demonstrations, and no explicit chain-of-thought exemplars are provided.
      \item The model is allowed to respond in free-form natural language; it may include reasoning text and then an answer, or simply output an answer, and there is no enforced answer template.
      \item A single sample is generated per question; there is no best-of-$n$ or multi-sample aggregation for this record.
    \end{itemize}

  \item \textbf{Other settings:}
    \begin{itemize}
      \item The system is used as a general-purpose large language model in its standard, non–reasoning-augmented mode, without special test-time ``extended thinking'' features or dedicated reasoning tokens.
      \item Decoding employs a moderate amount of stochasticity (nonzero temperature with nucleus sampling), consistent with settings used for non-reasoning baselines on related tasks.
      \item No retrieval augmentation, external tools, or auxiliary context sources (such as web search or separate knowledge bases) are used; the only information provided at inference time is the puzzle description and the question text.
      \item Context length requirements are minimal because each on-air item is short, and no special truncation rules are applied beyond the model’s default maximum output length.
    \end{itemize}
\end{itemize}
\medskip  
\medskip
\medskip
\textbf{Target metric}: 65 (accuracy)

\end{tcolorbox}

\newpage

\subsection{off-air setting description}

\definecolor{titleblue}{RGB}{38,116,186}
\definecolor{frameblue}{RGB}{38,116,186}
\definecolor{backblue}{RGB}{240,247,255}

\tcbset{
  mybox/.style={
    enhanced,
    breakable,
    colback=backblue,
    colframe=frameblue,
    boxrule=0.8pt,
    arc=3mm,
    left=10pt,right=10pt,top=10pt,bottom=10pt
  },
  mytitle/.style={
    fontupper=\bfseries\large\color{white},
    colback=titleblue,
    colframe=titleblue,
    boxrule=0pt,
    arc=2mm,
    left=8pt,right=8pt,top=6pt,bottom=6pt
  }
}

\lstdefinestyle{tightmono}{
  basicstyle=\ttfamily\small,
  columns=fullflexible,
  keepspaces=true,
  showstringspaces=false,
  breaklines=true,
  frame=single,
  framerule=0.3pt,
  rulecolor=\color{black!35},
  backgroundcolor=\color{white},
  tabsize=2
}

\setlist[itemize]{topsep=3pt,itemsep=2pt,parsep=0pt,leftmargin=1em,labelsep=0.5em}
\setlength{\parskip}{4pt}
\setlength{\parindent}{0pt}

\begin{tcolorbox}[mybox]

\textbf{Model:} GPT-4o

\begin{itemize}
  \item \textbf{Task:}
    \begin{itemize}
      \item Open-ended answer generation for verbal puzzles.
      \item Input: a short natural-language puzzle description specifying a pattern or constraint over words, letters, sounds, or common-knowledge facts.
      \item Answer space: effectively all possible strings in the language; solutions are typically a single word or short phrase, but can sometimes involve multiple words or multiple solution strings separated by punctuation.
      \item Evaluation target: check whether the model’s generated text contains the correct solution phrase(s) that satisfy the puzzle constraints.
    \end{itemize}

  \item \textbf{Data collection:}
    \begin{itemize}
      \item Derived from transcripts of a long-running weekly radio puzzle segment.
      \item Each transcript includes the previous week’s challenge puzzle and its official solution; these were scraped from publicly available web pages and then manually cleaned.
      \item Cleaning operations include making implicit temporal or geographic references explicit (for example, replacing relative time expressions like ``last year'' with the corresponding calendar year, or clarifying that a greeting comes from a country other than a reference country).
      \item Puzzles whose solutions admit very large numbers of valid answers (for example, cases where listeners submitted many distinct correct solutions) are removed because they are not well suited to automatic evaluation.
      \item Verbal explanations are stripped from the stored solutions so that the gold answer consists only of the target word(s), rather than a full explanatory sentence.
    \end{itemize}

  \item \textbf{Evaluation:}
    \begin{itemize}
      \item Metric: accuracy over individual puzzles.
      \item For each puzzle, the model is run once and produces a free-form completion that may include both reasoning and a final answer.
      \item Capitalization and punctuation are ignored during scoring; an output is judged correct if every phrase in the stored gold answer appears somewhere in the generated text.
      \item If the gold solution consists of multiple phrases, all of them must be present; otherwise the item is marked incorrect.
      \item No partial credit or ranking is used; each puzzle is counted as correct or incorrect, and mean accuracy over the full set of weekly challenges is reported.
    \end{itemize}

  \item \textbf{Difficulty characteristics:}
    \begin{itemize}
      \item Puzzles are designed for adult native speakers with broad general knowledge, and are intended to be sufficiently challenging that many listeners may think about them for days before the answer is revealed.
      \item Although they are hard to solve, solutions are straightforward for humans to verify once known, which also makes model errors easy to inspect.
      \item Many puzzles involve nontrivial wordplay such as letter rearrangements, constraints on letter counts, or sound-based relationships (rhymes, homophones), often combined with semantic or world-knowledge constraints (for example, cities, foods, well-known people, greetings, or other culturally specific entities).
      \item The collection spans multiple verbal-reasoning types, including letter-level manipulations, sound-based reasoning, visual or positional layouts of letters, riddles, and related trick-based formulations.
      \item Knowledge domains are broad, covering general knowledge, geography, history, popular culture, sports, politics, and arts; individual puzzles often require multi-step reasoning and candidate search rather than direct retrieval of a single fact.
    \end{itemize}

  \item \textbf{Subset:}
    \begin{itemize}
      \item This record corresponds to evaluation on the full collection of cleaned weekly challenge puzzles.
      \item No additional restrictions are applied by puzzle type, knowledge category, or time period beyond the cleaning and filtering described above.
    \end{itemize}
\end{itemize}

\begin{itemize}
  \item \textbf{Prompting strategy:}
    \begin{itemize}
      \item Zero-shot generation on the weekly challenge puzzles.
      \item For each puzzle, the input to the model consists solely of the puzzle text itself; there are no added instructions about answer format, no worked examples, and no few-shot demonstrations.
      \item The model is free to produce both a reasoning process (if any) and a final answer in an unconstrained natural-language completion.
      \item There is no explicit chain-of-thought prompting, and no special answer template is imposed; the system is treated as a standard text-completion model.
      \item Each reported accuracy is based on a single completion per puzzle, without best-of-$n$ sampling or aggregation.
    \end{itemize}

  \item \textbf{Other settings:}
    \begin{itemize}
      \item The evaluated model is a non-reasoning variant within a broader model family that also includes reasoning-augmented versions.
      \item It does not use extended test-time ``thinking'' modes or dedicated reasoning tokens; instead, it is run in its default completion mode.
      \item Decoding uses a relatively low sampling temperature and a nucleus-sampling cutoff (top-$p$), matching the configuration adopted for non-reasoning baselines on this benchmark.
      \item No external tools, retrieval components, or auxiliary context sources beyond the puzzle text are employed.
      \item Output length is not specially tuned beyond the default limits for this model class, and no special truncation rules are applied; each puzzle is handled as a single, short prompt with its corresponding single completion.
    \end{itemize}
\end{itemize}
\medskip  
\medskip
\medskip
\textbf{Target metric}: 6 (accuracy)

\end{tcolorbox}
\newpage

\section{\datasetname\ Samples}
\label{appendix:samples}

We randomly sample 5 instances from \datasetname\ for the demonstration.

\definecolor{titleblue}{RGB}{38,116,186}
\definecolor{frameblue}{RGB}{38,116,186}
\definecolor{backblue}{RGB}{240,247,255}

\tcbset{
  mybox/.style={
    enhanced,
    breakable,
    colback=backblue,
    colframe=frameblue,
    boxrule=0.8pt,
    arc=3mm,
    left=10pt,right=10pt,top=10pt,bottom=10pt
  },
  mytitle/.style={
    fontupper=\bfseries\large\color{white},
    colback=titleblue,
    colframe=titleblue,
    boxrule=0pt,
    arc=2mm,
    left=8pt,right=8pt,top=6pt,bottom=6pt
  }
}

\lstdefinestyle{tightmono}{
  basicstyle=\ttfamily\small,
  columns=fullflexible,
  keepspaces=true,
  showstringspaces=false,
  breaklines=true,
  frame=single,
  framerule=0.3pt,
  rulecolor=\color{black!35},
  backgroundcolor=\color{white},
  tabsize=2
}

\setlist[itemize]{topsep=3pt,itemsep=2pt,parsep=0pt,leftmargin=1em,labelsep=0.5em}
\setlength{\parskip}{4pt}
\setlength{\parindent}{0pt}

\begin{tcolorbox}[mybox]

\textbf{Source Papers}: result paper=$\texttt{2306.16638}$\\[0.6ex]
\textbf{Model}: GPT-4\\[0.6ex]
\textbf{Description}
\begin{itemize}
  \item \textbf{Task:}
    \begin{itemize}
      \item Binary classification of sentence pairs.
      \item Input: two short natural-language sentences presented as a pair.
      \item Label space: two classes indicating whether the second sentence logically negates the first (Yes/No; encoded as a binary output).
      \item Evaluation target: identify logical negation between sentences rather than relying on surface-level cues.
    \end{itemize}

  \item \textbf{Data collection:}
    \begin{itemize}
      \item Custom corpus created by the author using designed sentence templates.
      \item For each pair, components were systematically negated (or left unmodified) to generate balanced positive and negative examples.
      \item Construction followed classical logic principles to ensure ``negation'' reflects logical contradiction, with emphasis on structures prone to errors (e.g., conditionals).
      \item Sentence length: approximately 5--20 words, with a roughly normal distribution.
      \item Domains: diverse everyday and professional contexts; pairs span multiple topics.
    \end{itemize}

  \item \textbf{Evaluation:}
    \begin{itemize}
      \item Metric: Recall calculated for the positive class (pairs where the second sentence negates the first), comparing model predictions with gold-standard labels across the entire dataset.
    \end{itemize}

  \item \textbf{Difficulty characteristics:}
    \begin{itemize}
      \item Balanced two-class classification task.
      \item Inclusion of logically complex constructions (e.g., conditional statements), rendering surface-level ``not'' cues unreliable.
      \item Short but variable sentence lengths (5--20 words).
      \item Designed to assess logical reasoning robustness across diverse contexts.
    \end{itemize}
\end{itemize}

\begin{itemize}
  \item \textbf{Prompting strategy:}
    \begin{itemize}
      \item Zero-shot classification.
      \item A fixed instruction presents the two sentences and asks whether the second negates the first, requiring a binary response (Yes/No encoded as 1/0).
      \item No in-context examples, chain-of-thought reasoning, or auxiliary decoding constraints used.
    \end{itemize}

  \item \textbf{Other settings:}
    \begin{itemize}
      \item Evaluated via an API for the later-generation model in this family; each pair assessed once using the fixed prompt.
      \item No retrieval augmentation, external tools, or special truncation/windowing rules reported.
    \end{itemize}
\end{itemize}
\medskip  
\medskip
\medskip
\textbf{Target metric}: 72.8 (recall)

\end{tcolorbox}

\begin{tcolorbox}[mybox]

\textbf{Source Papers}: result paper=\texttt{2403.00092}\\[0.6ex]
\textbf{Model}: GPT-4\\[0.6ex]
\textbf{Description}
\begin{itemize}
  \item \textbf{Task:}
    \begin{itemize}
      \item Predict symbolic planning domain actions from natural-language procedural text.
      \item Input: a natural-language procedure together with a domain header listing available types, predicates, and the names of actions to be filled in.
      \item Output: a domain file’s action definitions; each action must include typed parameters, a conjunction of preconditions, and a conjunction of effects, expressed in a formal planning-language syntax.
      \item Constraint: models must not add, remove, or rename actions.
      \item Evaluation target: reproduce the gold action definitions so that the predicted domain file matches the reference.
    \end{itemize}

  \item \textbf{Data collection:}
    \begin{itemize}
      \item Source texts are open-domain how-to guides from the web covering varied everyday and survival-style topics.
      \item For each procedure, trained experts manually created paired symbolic planning representations: a domain file (types, predicates, actions) and associated problem files (entities, initial and goal states).
      \item Representations are planner-ready and not tied to any specific simulator.
    \end{itemize}

  \item \textbf{Evaluation:}
    \begin{itemize}
      \item Metric: action-level accuracy under intrinsic evaluation.
      \item Protocol: compare predicted actions with the reference actions in the gold domain file using a semantic-equivalence check that disregards variable renaming and conjunct order; parameters must match by type, and predicates must align in arity and argument mapping.
    \end{itemize}

  \item \textbf{Subset:}
   \begin{itemize}
    \item intrinsic—only direct semantic alignment of predicted action definitions is assessed; no plan execution/solvability is considered.
   \end{itemize}

  \item \textbf{Difficulty characteristics:}
    \begin{itemize}
      \item Open-domain content with diverse topics.
      \item Requires extracting relevant information, inferring implicit entity states, and precisely translating into a low-resource, domain-specific formal language.
      \item Input texts can be long (from concise summaries to multi-paragraph procedures), creating context-length and information-extraction challenges.
    \end{itemize}
\end{itemize}

\begin{itemize}
  \item \textbf{Prompting strategy:}
    \begin{itemize}
      \item Zero-shot direct instruction without step-by-step reasoning scaffolding.
      \item The prompt provides the domain header (types, predicates, action names) and the relevant procedure text, and asks the model to complete each action’s parameters, preconditions, and effects in the required formal syntax while preserving the given action set.
    \end{itemize}

  \item \textbf{Other settings:}
    \begin{itemize}
      \item Evaluated via an API for GPT\text{-}4; the input budget was capped at roughly ten thousand tokens for these runs.
    \end{itemize}
\end{itemize}
\medskip
\medskip
\medskip
\textbf{Target metric}: 15.9 (accuracy)

\end{tcolorbox}

\newpage

\begin{tcolorbox}[mybox]

\textbf{Source Papers}: result paper=\texttt{2310.11191}, dataset paper=\texttt{2104.05767}\\[0.6ex]
\textbf{Model}: GPT-4\\[0.6ex]
\textbf{Description}
\begin{itemize}
  \item \textbf{Task:}
    \begin{itemize}
      \item Paragraph-to-document–level text simplification in the medical domain.
      \item Input: one or more paragraphs from technical abstracts of clinical evidence reviews.
      \item Output: a plain-language, open-ended rewrite that retains key findings.
      \item Answer space: unrestricted natural language (no predefined options).
      \item Evaluation target: alignment of the generated simplification with the reference plain-language version in content and lexical overlap.
    \end{itemize}

  \item \textbf{Data collection:}
    \begin{itemize}
      \item Source texts: technical abstracts from evidence syntheses on clinical topics; targets: plain-language summaries authored by domain experts for non-expert audiences.
      \item English-only; sourced from a large online repository of medical systematic reviews.
      \item Abstracts and corresponding lay summaries were paired via heuristic section matching: results-focused abstract sections were retained; for lay summaries, sections discussing studies/results (or the first paragraph containing terms like ``study/trial'' for unstructured summaries) were selected. Pairs with large length discrepancies were removed; examples were filtered to fit standard transformer context budgets.
      \item Note: lay summaries are not direct simplifications of abstracts (derived from full reviews), yielding partial content overlap rather than sentence-level parallelism.
    \end{itemize}

  \item \textbf{Evaluation:}
    \begin{itemize}
      \item Metric: ROUGE-LSum (F1, longest common subsequence at the summary level) between model output and the single reference lay summary.
    \end{itemize}

  \item \textbf{Difficulty characteristics:}
    \begin{itemize}
      \item Domain shift and style gap: targets are stylistically simpler but only marginally easier on readability metrics; success requires controlled paraphrasing and style adaptation beyond sentence/word shortening.
      \item Technical jargon and statistical elements (e.g., measurements, confidence intervals) increase lexical/semantic complexity.
      \item Non-parallelity: targets may include information absent from abstracts (e.g., currency statements), creating source–target mismatches and risk of hallucination.
      \item Long, multi-paragraph inputs necessitate document-level content selection and abstraction.
    \end{itemize}
\end{itemize}

\begin{itemize}
  \item \textbf{Prompting strategy:}
    \begin{itemize}
      \item Zero-shot instruction prompting: a system instruction specifies a simplification assistant role; a user instruction asks to simplify the provided text. No examples or chain-of-thought demonstrations are provided.
    \end{itemize}

  \item \textbf{Other settings:}
    \begin{itemize}
      \item No retrieval augmentation, external tools, or special decoding constraints reported for this record.
    \end{itemize}
\end{itemize}
\medskip
\medskip
\medskip
\textbf{Target metric}: 33.9 (rouge)

\end{tcolorbox}

\newpage

\begin{tcolorbox}[mybox]
\textbf{Source Papers}: result paper=\texttt{2310.20105}\\[0.6ex]
\textbf{Model}: GPT-4\\[0.6ex]
\textbf{Description}
\begin{itemize}
  \item \textbf{Task:}
    \begin{itemize}
      \item Single-label, multi-class text classification of student help requests from introductory programming courses.
      \item Input: a natural-language issue description, optionally including a code snippet and an error message; the prompt also provides a numeric indicator of alignment between the message and assignment instructions.
      \item Label space: six mutually exclusive categories—three debugging subcategories (error message present; expected outcome specified; both present), plus implementation, understanding, and a nothing/insufficient-content class.
      \item Target: assign a single category to each request.
    \end{itemize}

  \item \textbf{Data collection:}
    \begin{itemize}
      \item Source: requests submitted via a course help assistant using a semi-structured form.
      \item Captured contents: issue description, optional code and error message (programming language collected by the form but not used in the prompt).
      \item Annotation: two human coders independently labeled requests; disagreements were resolved by adopting the more experienced coder’s label; inter-rater agreement reported as substantial.
      \item Domain/language/licensing: programming-education context; language unspecified; no licensing details provided.
    \end{itemize}

  \item \textbf{Evaluation:}
    \begin{itemize}
      \item Metric: F1 score for the specific category (per-class F1) on the test set, comparing model predictions with human reference labels.
    \end{itemize}

 \item \textbf{Subset:}
   \begin{itemize} 
    \item Implementation category only—requests seeking guidance on implementing code for a specific assignment problem (often include code and/or explicit references to instructions).
   \end{itemize}
   
  \item \textbf{Difficulty characteristics:}
    \begin{itemize}
      \item Six-way classification with nuanced distinctions based on subtle cues (e.g., presence/absence of code, explicit error messages vs.\ desired outcomes, references to instructions).
      \item Inputs may include code and error traces; no prompt-length constraints were reported to affect evaluation.
    \end{itemize}
\end{itemize}

\begin{itemize}
  \item \textbf{Prompting strategy:}
    \begin{itemize}
      \item Zero-shot classification (no in-context examples, no chain-of-thogught).
      \item System instruction outlines the full coding scheme and decision criteria (adapted from human-coder guidelines) and requires output of a single category label without explanation.
      \item User message includes: the student’s issue description, any provided code, any provided error message, and a numeric “issue–instructions equivalence” percentage indicating overlap with assignment instructions (\% only used as context).
    \end{itemize}

  \item \textbf{Other settings:}
    \begin{itemize}
      \item Inference via API, one request at a time.
      \item temperature 0.0, top\_p 1, max\_tokens 10, frequency\_penalty 0, presence\_penalty 0.
    \end{itemize}
\end{itemize}
\medskip
\medskip
\medskip
\textbf{Target metric}: 85 (F1)

\end{tcolorbox}

\newpage

\begin{tcolorbox}[mybox]

\textbf{Source Papers}: result paper=\texttt{2406.01238}, dataset paper=\texttt{1803.06643}\\[0.6ex]
\textbf{Model}: GPT-4\\[0.6ex]
\textbf{Description}
\begin{itemize}
  \item \textbf{Task:}
    \begin{itemize}
      \item Open-domain natural-language question answering over knowledge graphs.
      \item Input: a single English question; the system may decompose it into sub-questions and navigate a knowledge graph.
      \item Output: one or more target entities/values from the graph; the final prediction is the top candidate.
      \item Evaluation target: whether the top predicted answer matches any reference answer.
    \end{itemize}

  \item \textbf{Data collection:}
    \begin{itemize}
      \item Built by generating complex queries from a simpler QA corpus that pairs questions with graph queries over a large collaborative knowledge graph.
      \item Complexity introduced via function composition, conjunctions, superlatives, and comparatives.
      \item Machine-generated templates derived from graph predicates were paraphrased by crowd workers into natural language; gold answers obtained by executing graph queries.
      \item Domain: broad, open-domain factual knowledge; Language: English; Provenance: curated knowledge graph + crowdsourced paraphrases.
    \end{itemize}

  \item \textbf{Evaluation:}
    \begin{itemize}
      \item Metric: answer accuracy (Hits@1), i.e., whether the top prediction matches any gold answer string/entity.
    \end{itemize}

  \item \textbf{Difficulty characteristics:}
    \begin{itemize}
      \item Multi-step reasoning with compositions and set operations (e.g., conjunctions), plus superlative/comparative constraints.
      \item Paraphrase variability: human rewrites introduce synonymy, reordering, omissions, and additions.
      \item Open-domain coverage and entity-centric reasoning increase ambiguity, requiring precise entity typing and relation selection.
    \end{itemize}
\end{itemize}

\begin{itemize}
  \item \textbf{Prompting strategy:}
    \begin{itemize}
      \item Strategic, iterative LLM-driven planning over a knowledge graph (EffiQA). The high-capacity LLM performs: 1) global planning to decompose the question and generate exploration instructions (with simulated answers to guide search); 2) delegated efficient KG exploration via breadth-first expansion with semantic pruning using fine-grained entity typing and similarity to simulated answers; 3) self-reflection to review partial results, revise the plan, and iterate until sufficient evidence is gathered.
    \end{itemize}

  \item \textbf{Other settings:}
    \begin{itemize}
      \item LLM temperature set to 0 for deterministic outputs.
      \item Exploration uses breadth-first search with thresholds to control branching; relations/entities not matching planned types/qualifiers are pruned.
      \item A plug-in encoder-based classifier (fine-grained entity typing) supports constrained semantic matching to maintain recall while reducing search space.
      \item Stopping criterion: output an answer when results are satisfactory; otherwise replan and iterate.
    \end{itemize}
\end{itemize}
\medskip
\medskip
\medskip
\textbf{Target metric}: 69.5 (accuracy)

\end{tcolorbox}
\newpage 

\section{Prompts}

\subsection{Prompt for prediction using search}

\definecolor{titleblue}{RGB}{96,96,96}
\definecolor{frameblue}{RGB}{190,190,190} 
\definecolor{backblue}{RGB}{248,248,248}  

\tcbset{
  mybox/.style={
    enhanced,
    breakable,
    colback=backblue,
    colframe=frameblue,
    boxrule=0.8pt,
    arc=3mm,
    left=10pt,right=10pt,top=10pt,bottom=10pt
  },
  mytitle/.style={
    fontupper=\bfseries\large\color{white},
    colback=titleblue,
    colframe=titleblue,
    boxrule=0pt,
    arc=2mm,
    left=8pt,right=8pt,top=6pt,bottom=6pt
  }
}

\lstdefinestyle{tightmono}{
  basicstyle=\ttfamily\small,
  columns=fullflexible,
  keepspaces=true,
  showstringspaces=false,
  breaklines=true,
  frame=single,
  framerule=0.3pt,
  rulecolor=\color{black!35},
  backgroundcolor=\color{white},
  tabsize=2
}

\setlist[itemize]{topsep=3pt,itemsep=2pt,parsep=0pt,leftmargin=1em,labelsep=0.5em}
\setlength{\parskip}{4pt}
\setlength{\parindent}{0pt}

\begin{tcolorbox}[mybox]

\textbf{Prompt}\\[0.6ex]

\medskip

Given an experimental record (task, model, and setup), search for supporting evidence and estimate the LLM’s performance for the specified setup.\\[0.6ex]

\textbf{Inputs}\\[0.2ex]
\begin{itemize}
  \item Target model: model evaluated on.
  \item Metric family: metric family used to evaluate the experimental record.
  \item Redacted description: description of the experimental record (e.g., task, dataset collection, evaluation metric, experimental setup).
\end{itemize}

\medskip  
\medskip

\textbf{Search Process (REQUIRED)}\\[0.2ex]
\begin{itemize}
  \item Use the search tool up to four times to gather evidence.
  \item Keep queries concise; the retrieval module cannot handle complex queries.
  \item Focus queries on the Target model evaluated on related datasets/tasks (e.g., GPT-4o clinical summarization rouge).
  \item Submit at most one query per turn; after each query, the tool returns matching paper PDFs.
  \item Plan what to search before calling the tool.
  \item If retrieved evidence is irrelevant, do not reference it for the prediction.
\end{itemize}

\medskip

\textbf{Confidence Assessment}\\[0.2ex]
\begin{itemize}
  \item After producing a performance prediction, critically review your reasoning and the evidence considered.
  \item Classify confidence into one of:
    \begin{itemize}
      \item Almost no chance, Highly unlikely, Chances are slight, Unlikely, Less than even,
      Better than even, Likely, Very good chance, Highly likely, Almost certain.
    \end{itemize}
\end{itemize}

\medskip

\textbf{Output Format}\\[0.2ex]
\begin{enumerate}
  \item Prediction: \textbf{Point estimate} on a 0--100 scale as \textbackslash boxed\{\}.
  \item Rationale: brief, ordered reasons mapping features \(\rightarrow\) expected performance shifts (e.g., 5 options (20\% baseline) + few-shot CoT (+) + domain shift (--) + exact match (--) → net expectation …).
  \item Confidence Assessment: one class name wrapped in \textbackslash confidence\{\}.
  \item Retrieval Evidence: list search queries, retrieved papers, and concise summaries of relevant findings from those papers.
\end{enumerate}

\end{tcolorbox}

\newpage

\subsection{Prompt for prediction without using search}

\begin{tcolorbox}[mybox]

\textbf{Prompt}\\[0.6ex]

\medskip

Given an experimental record (task, model, and setup), estimate the LLM’s performance for the specified setup.\\[0.6ex]

\textbf{Inputs}\\[0.2ex]
\begin{itemize}
  \item Target model: model evaluated on.
  \item Metric family: metric family used to evaluate the experimental record.
  \item Redacted description: description of the experimental record (e.g., task, dataset collection, evaluation metric, experimental setup).
\end{itemize}

\medskip  
\medskip

\textbf{Confidence Assessment}\\[0.2ex]
\begin{itemize}
  \item After producing a performance prediction, critically review your reasoning and the evidence considered.
  \item Classify confidence into one of:
    \begin{itemize}
      \item Almost no chance, Highly unlikely, Chances are slight, Unlikely, Less than even,
      Better than even, Likely, Very good chance, Highly likely, Almost certain.
    \end{itemize}
\end{itemize}

\medskip

\textbf{Output Format}\\[0.2ex]
\begin{enumerate}
  \item Prediction: \textbf{Point estimate} on a 0--100 scale as \textbackslash boxed\{\}.
  \item Rationale: brief, ordered reasons mapping features \(\rightarrow\) expected performance shifts (e.g., 5 options (20\% baseline) + few-shot CoT (+) + domain shift (--) + exact match (--) → net expectation …).
  \item Confidence Assessment: one class name wrapped in \textbackslash confidence\{\}.
\end{enumerate}

\end{tcolorbox}

\newpage

\section{Compliance with arXiv Access Policies}
\label{app:arxiv-compliance}

We accessed arXiv content in accordance with arXiv’s published guidance for automated use, including the
\texttt{robots.txt}\footnote{\url{https://arxiv.org/robots.txt}} and the “Robots Beware” help page.\footnote{\url{https://info.arxiv.org/help/robots.html}}
Our goal was to minimize load on arXiv’s infrastructure while ensuring traceable, policy-compliant retrieval.

\paragraph{Streaming evaluation.}
For the prospective streaming evaluation, we collected newly released papers from \texttt{cs.CL}, \texttt{cs.LG}, and \texttt{cs.AI} between September 5 and September 23, 2025, using arXiv’s official programmatic interfaces. 
We obtained 79 eligible experimental records from 52 papers and issued predictions before third-party search indexing. Records were processed using the same extraction and filtering pipeline as \datasetname.

\paragraph{Scope of access.}
Agents fetched only URLs permitted by \texttt{robots.txt} and avoided disallowed paths. We did not perform indiscriminate crawling or attempt corpus-scale downloads via the human-facing site.

\paragraph{Preferred interfaces.}
Wherever possible, we used arXiv’s programmatic interfaces (e.g., API/OAI-PMH/RSS) instead of scraping HTML pages, consistent with the help page’s guidance to favor machine interfaces for automated access.\footnotemark[2]

\paragraph{Robots rules and rate limiting.}
Our clients honored \texttt{robots.txt} directives (allow/disallow and crawl-delay) and enforced conservative global throttling with backoff on errors. We limited parallelism and maintained a single logical connection per agent run.

\paragraph{Identification and advance notice.}
All requests included a descriptive \texttt{User-Agent} string and a monitored contact email, as recommended.\footnotemark[2]
Prior to large-scale experiments, we emailed the arXiv team responsible for automated access to describe our usage pattern and confirm that no special coordination was required.

\paragraph{Low-impact retrieval policy.}
To further reduce load, each retrieval-capable agent capped (i) the number of queries it could issue and (ii) the number of documents returned per query. We cached responses, deduplicated URLs, and normalized links to canonical arXiv endpoints to avoid redundant traffic.

\paragraph{Monitoring and fail-safe behavior.}
We continuously monitored HTTP status codes and immediately slowed or halted traffic upon receiving \texttt{429}/\texttt{403}/\texttt{5xx} responses. Our clients also rechecked \texttt{robots.txt} at regular intervals to respect policy changes; any change that tightened restrictions triggered an automatic stop.

\paragraph{Summary.}
In aggregate, our practices align with arXiv’s automated access rules: we honored \texttt{robots.txt}, preferred official machine interfaces, rate-limited conservatively, identified ourselves, notified arXiv in advance, and implemented additional safeguards to minimize load.

\end{document}